\documentclass[11pt, letterpaper]{cmu}

\usepackage[comma,numbers,sort,compress]{natbib}
\usepackage{microtype}
\usepackage{float}
\usepackage{mathtools}
\usepackage{mathrsfs}
\usepackage{nicefrac}
\usepackage{dsfont}
\usepackage{bbm}
\usepackage{multirow}
\usepackage[export]{adjustbox}
\usepackage{wrapfig}
\usepackage{subcaption}
\usepackage{dblfloatfix}
\usepackage{listings}
\usepackage{setspace}
\usepackage{algpseudocode}
\usepackage{algorithm}
\usepackage{array}
\usepackage{makecell}
\usepackage{lineno}
\usepackage{soul}
\usepackage[most,skins,theorems]{tcolorbox}
\usepackage[textsize=tiny]{todonotes}
\usepackage[capitalize,noabbrev]{cleveref}
\usepackage{capt-of}  
\usepackage{etoolbox}

\setboolean{logo}{true}

\hypersetup{
    colorlinks = true,
    citecolor = {magenta},
    linkcolor = {blue},
    urlcolor = {blue},
}

\usepackage{amsthm}
\newtheorem{lemma}{Lemma}
\newtheorem{remark}{Remark}

\newtheorem{assumption}{Assumption}
\newtheorem{definition}{Definition}
\newtheorem{corollary}{Corollary}

\definecolor{highlightmistake}{RGB}{255, 179, 179}
\definecolor{highlightcorrect}{RGB}{179, 255, 179}
\definecolor{rliableolive}{HTML}{BBCC33}
\definecolor{rliableblue}{HTML}{77AADD}
\definecolor{rliablered}{HTML}{EE8866}
\definecolor{myblue}{rgb}{0.1,0.4,0.9}
\definecolor{lightblue}{rgb}{0.22,0.45,0.70}
\definecolor{violet}{HTML}{6A0DAD}
\definecolor{darkblue}{rgb}{0, 0, 0.5}

\newcolumntype{Y}{>{\centering\arraybackslash}X}
\newcolumntype{L}[1]{>{\raggedright\arraybackslash}p{#1}}

\newtcolorbox{promptbox}[2][]{
listing only,
enhanced,
breakable,
colback=rliableolive!13!white,
colframe=black,
fontupper=\ttfamily,
title=#2,
#1}

\tcbset{
  aibox/.style={
    width=\linewidth,
    top=8pt,
    bottom=4pt,
    colback=rliableolive!8!white,
    colframe=black,
    colbacktitle=black,
    enhanced,
    center,
    attach boxed title to top left={yshift=-0.1in,xshift=0.15in},
    boxed title style={boxrule=0pt,colframe=white,},
  }
}
\newtcolorbox{AIbox}[2][]{aibox,title=#2,#1}

\newcommand{\ourtitle}{Rethinking Supervision Scaling in On-Policy Distillation: From Dense Supervision to Stochastic Horizons}

\usepackage{amsmath,amsfonts,bm}

\def\eqref#1{Eq.~\ref{#1}}

\def\1{\bm{1}}

\DeclareMathAlphabet{\mathsfit}{\encodingdefault}{\sfdefault}{m}{sl}
\SetMathAlphabet{\mathsfit}{bold}{\encodingdefault}{\sfdefault}{bx}{n}

\title{\ourtitle}
\author[1]{Xinhao Yao}
\author[1]{Zhenxing Dou}
\author[1]{Yong Liu}

\affil[1]{Renmin University of China}

\correspondingauthor{liuyonggsai@ruc.edu.cn}

\providecommand{\theabstract}{}

\begin{document}
\maketitle

\vspace{-1.5cm}
\begin{center}
\begin{minipage}{0.94\textwidth}
  \centering
  \includegraphics[width=0.90\textwidth]{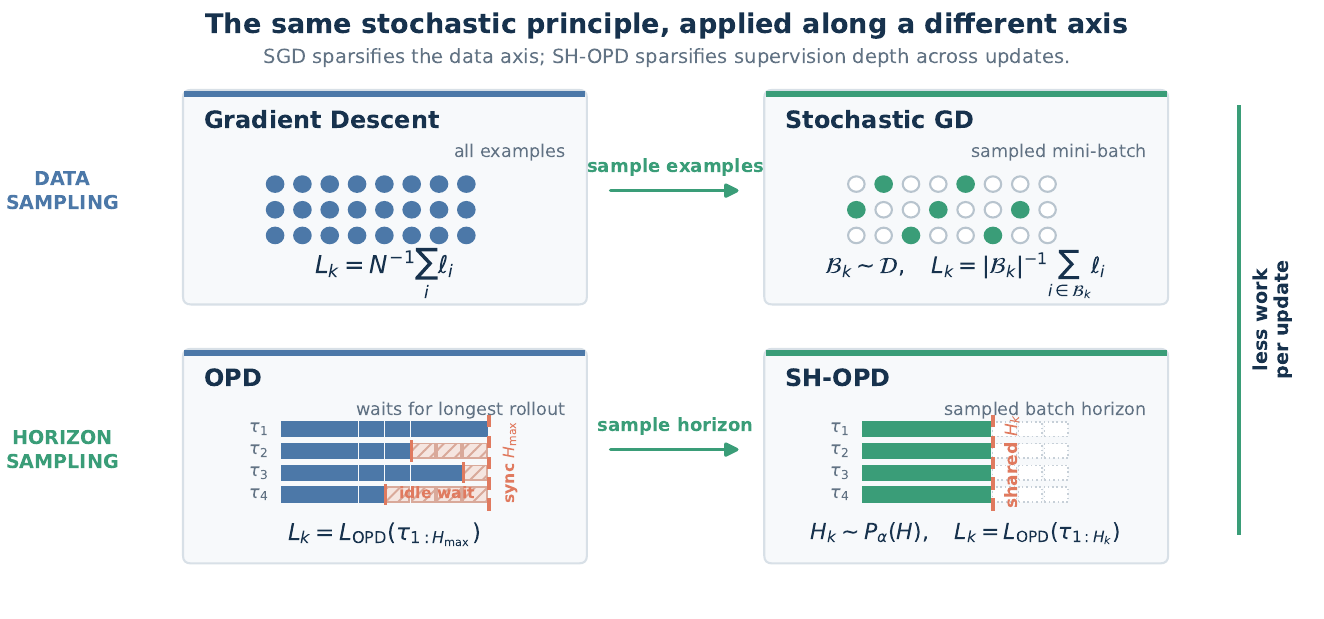}
  \vspace{-0.15cm}
\captionof{figure}{\footnotesize{\textbf{Batch-level stochastic horizon sampling in SH-OPD.}}
Just as mini-batch SGD reduces the per-update computation of full-batch gradient descent (GD) by sampling training examples, SH-OPD reduces the rollout cost of standard OPD by sampling a shared horizon for each batch. All trajectories are generated in parallel, with each response terminating at EOS or the shared horizon \(H_t\). By frequently sampling shorter horizons, SH-OPD reduces the computational overhead of repeatedly reinforcing overlapping supervision, while stochastic resampling across updates preserves access to potentially complementary supervision at deeper response positions. All generated tokens are optimized using the standard token-mean OPD objective.}
\label{fig:sgd_shopd_analogy}
\end{minipage}
\end{center}
\vspace{-0.35cm}

\noindent\textbf{Abstract: }
On-policy distillation (OPD) provides dense token-level supervision along student-generated trajectories, yet the learning benefits of longer trajectories may not scale proportionally with their computational cost. We investigate how supervision utility evolves with trajectory length. Controlled experiments reveal diminishing returns from additional supervised tokens, while showing that later positions remain learnable and exhibit substantial cross-position supervision transfer. Gradient analysis further identifies shared update components across response positions and diverse position-specific residuals whose cross-rollout consistency decreases at greater depths. Motivated by these findings, we propose \textbf{Stochastic Horizons On-Policy Distillation (SH-OPD)}, which reduces redundant supervision by randomly varying rollout horizons while preserving access to deeper response states. SH-OPD samples a shared horizon for each training batch, enabling efficient parallel generation without per-sequence horizon scheduling overhead. We characterize the resulting reweighting bias and derive a compute–reweighting trade-off under standard optimization assumptions. Experiments on mathematical reasoning and code generation demonstrate that SH-OPD achieves slightly higher mathematical reasoning accuracy at approximately 45\% of the per-update cost of standard OPD, while maintaining competitive code generation performance at around 60\% of the baseline cost. These results highlight the benefits of efficient supervision allocation across response depths.

\vspace{0.3em}
\noindent\textbf{Code:} \url{https://github.com/chen123CtrlS/SH-OPD}
\vspace{0.3em}

\vspace{-0.4cm}
\section{Introduction}
\vspace{-0.2cm}

On-policy distillation (OPD) has emerged as an effective paradigm for language model post-training \citep{agarwal2024onpolicy,gu2024minillm,kimik3,glm5,nvidia2026nemotron35}. Its core principle is to generate trajectories using the student policy and provide dense token-level supervision from a stronger teacher at each visited state. By supervising student-generated trajectories, OPD mitigates the train–inference distribution mismatch inherent in off-policy distillation. Unlike outcome-based reinforcement learning, which relies primarily on sequence-level feedback, OPD provides \(O(N)\) token-level supervision \citep{lu2025onpolicydistillation} opportunities along a trajectory of length \(N\), contributing to its remarkable data efficiency \citep{li2026rethinking,fu2026oneshotopd}. However, longer trajectories incur greater costs in generation, teacher evaluation, and student optimization. More fundamentally, more supervised tokens do not necessarily yield proportionally more useful learning signals. This raises a central question: \textit{\textbf{How does effective supervision in OPD scale with trajectory length?}}

Recent studies suggest that full-trajectory supervision is not always necessary for effective OPD. Prefix distillation and horizon-control methods achieve competitive performance with substantially shorter or progressively extended rollouts \citep{zhang-etal-2026-fast,popd_topd}. Position-aware analyses attribute the reduced utility of later tokens to accumulated teacher–student discrepancies \citep{xie2026positionbiasonpolicydistillation}, while token-selection methods show that a small subset of informative tokens can preserve much of the benefit of dense supervision \citep{xu2026tiptokenimportanceonpolicy}. Beyond supervision allocation, \citet{fu2026oneshotopd} investigate state-space coverage and teacher–student alignment, while \citet{shen2026geometryonpolicydistillation} and \citet{yu2026densesupervisionsparseupdates} characterize the geometric structure of OPD parameter updates. Together, these studies reveal the non-uniform benefits of token-level supervision and the structured nature of OPD updates. However, \textbf{why additional supervised tokens yield diminishing returns remains insufficiently understood}. In particular, supervision from different trajectory positions may induce overlapping parameter updates, limiting the marginal benefits of extending the supervision horizon. Meanwhile, position-specific gradients may capture additional information that varies across sampled trajectories. These possibilities motivate a closer examination of how supervision interacts across trajectory positions and whether additional tokens provide genuinely new learning signals.

To investigate these diminishing returns, we examine how supervision utility changes with trajectory length and position (Sections \ref{subsec:fixed_prefix} and \ref{subsec:cross-block}). Fixed-prefix replay experiments show that increasing the number of supervised tokens yields progressively smaller gains in downstream performance. However, equal-budget position controls reveal that later trajectory blocks remain highly learnable, and supervision on one block substantially improves teacher–student agreement in other, unsupervised blocks. These findings suggest that \textbf{diminishing returns arise, at least in part, from substantial supervision overlap across trajectory positions, rather than an inherent lack of useful signals in later tokens}.

To explain this cross-position transfer, we examine the geometry of blockwise OPD gradients (Section \ref{subsec:geometry}). We find substantial gradient alignment across trajectory positions, indicating a shared update component. Beyond this shared component, position-specific residuals exhibit diverse directions but become less consistent across sampled trajectories as response depth increases. These findings suggest that \textbf{longer trajectories repeatedly reinforce shared updates while introducing increasingly trajectory-dependent variations}. This provides an explanation for the diminishing returns of dense supervision and motivates reducing redundant supervision while preserving access to deeper response states.

Building on these findings, we propose \textbf{Stochastic Horizons On-Policy Distillation (SH-OPD)}, which reduces redundant supervision while preserving access to deeper response states (Section \ref{main:sec_shopd}). SH-OPD samples a batch-level rollout horizon from a power-survival distribution, favoring shorter trajectories while occasionally exploring longer ones. We further analyze the resulting horizon-induced reweighting bias through the Gram matrix of expected depth-specific gradient residuals. Under standard optimization assumptions, our theoretical analysis characterizes the trade-off between compute-adjusted stochastic optimization error and the distortion of stable depth-specific supervision. Experiments on mathematical reasoning and code generation demonstrate that SH-OPD substantially reduces training cost while maintaining competitive performance. On mathematical reasoning, SH-OPD achieves slightly higher accuracy than standard OPD while using less than half the per-update token-compute and wall-time costs. On code generation, it maintains competitive performance with approximately 60\% of the baseline cost. These results highlight the benefits of allocating supervision more efficiently across response depths.

Our main contributions are summarized as follows:
\begin{itemize}
    \item \textbf{Supervision scaling.} We reveal diminishing returns from increasing supervision length and demonstrate substantial cross-position transfer, showing that additional supervised tokens do not necessarily provide proportional learning benefits.

    \item \textbf{Gradient geometry and theoretical analysis.} We characterize shared and position-specific gradient components and their cross-rollout consistency. Building on this geometric structure, we analyze horizon-induced reweighting bias and establish a compute–reweighting trade-off under standard optimization assumptions.

    \item \textbf{Efficient method.} We propose SH-OPD, a simple stochastic horizon strategy that reduces redundant supervision while preserving access to deeper response states, achieving substantial computational savings with competitive performance across tasks. 
\end{itemize}

\vspace{-0.3cm}
\section{Preliminaries, Definitions, and Notation}
\label{sec:prelim}
\vspace{-0.2cm}

\textbf{On-Policy Distillation.} Let $x \sim \mathcal{D}$ denote an input prompt, $\pi_\theta$ the student policy parameterized by $\theta$, and $\pi_{\mathrm{t}}$ a fixed teacher policy. Given $x$, the student generates a response
$y = (y_1,\ldots,y_N)\sim \pi_\theta(\cdot \mid x)$, where $N := |y|$ denotes the number of response tokens. We use $s_i := (x,y_{<i})$ to denote the autoregressive state before generating token $y_i$. On-Policy Distillation \citep{agarwal2024onpolicy,gu2024minillm} (OPD) minimizes the reverse KL divergence between the student and teacher distributions evaluated on prefixes sampled from the student policy:
$\mathcal{K}(\theta)
=
\mathbb{E}_{x\sim\mathcal{D},\,
y\sim\pi_\theta(\cdot\mid x)}
\left[
\sum_{i=1}^{N}
\left(
\log \pi_\theta(y_i\mid s_i)
-
\log \pi_{\mathrm{t}}(y_i\mid s_i)
\right)
\right]$. For each student-sampled token, OPD defines a token-level teacher preference signal
$A_i
:=
\text{stopgrad}(\log
\frac{
\pi_{\mathrm{t}}(y_i\mid s_i)
}{
\pi_\theta(y_i\mid s_i)
})$. In practice, the corresponding policy-gradient-style \citep{yang2026learningteachergeneralizedonpolicy} OPD surrogate can be written as:

\begin{equation}
\label{eq:opd-loss}
\mathcal{L}_{\mathrm{OPD}}
(\theta)
= \mathbb{E}_{x\sim\mathcal{D},\,
y\sim\pi_\theta(\cdot\mid x)}
\left[
-\frac{1}{N}
\sum_{i=1}^{N}
A_i
\log
\pi_\theta(y_i\mid s_i) \right ].
\end{equation}

The loss function (\eqref{eq:opd-loss}) assigns positive or negative token-level credit according to the teacher's relative preference, thereby steering the student toward the teacher's behavior at each state. For convenience, define the per-token OPD loss as
$
\ell_i(\theta;x,y)
:=
-
A_i
\log
\pi_\theta(y_i\mid s_i)
$,
such that
\begin{equation}
\label{eq:full-token-loss}
\mathcal{L}_{\mathrm{OPD}}
(\theta)
=
\mathbb{E}_{x\sim\mathcal{D},\,
y\sim\pi_\theta(\cdot\mid x)}
\left[
\frac{1}{N}
\sum_{i=1}^{N}
\ell_i(\theta;x,y)
\right ].
\end{equation}

\textbf{Blockwise Decomposition.}
To examine how the utility of dense supervision evolves along a response, we partition each response into contiguous token blocks.
Let $m$ denote the target block size and define
$K
=
\left\lceil
\frac{N}{m}
\right\rceil$
as the number of blocks.
The token indices belonging to the $k$-th block are
$
\mathcal{I}_k
=
\left\{
(k-1)m+1,\ldots,\min(km,N)
\right\},
k=1,\ldots,K.
$
Let
$
    n_k := |\mathcal{I}_k|
$
denote the number of tokens contained in block $k$. We define the average OPD loss within block $k$ as
$
\mathcal{L}_k(\theta;x,y)
:=
\frac{1}{n_k}
\sum_{i\in\mathcal{I}_k}
\ell_i(\theta;x,y)$. The full-response OPD loss admits the exact decomposition
$
\widehat{\mathcal{L}}_{\mathrm{OPD}}
(\theta;x,y)
=
\sum_{k=1}^{K}
\frac{n_k}{N}
\mathcal{L}_k(\theta;x,y)
.
$ When all blocks have equal size, i.e.,
$n_k=m$ and $N=Km$, ~\eqref{eq:full-token-loss} reduces to
\begin{equation}
\label{eq:equal-block-decomposition}
\widehat{\mathcal{L}}_{\mathrm{OPD}}
(\theta)
=
\mathbb{E}_{x\sim\mathcal{D},\,
y\sim\pi_\theta(\cdot\mid x)}
\left[\frac{1}{K}
\sum_{k=1}^{K}
\mathcal{L}_k(\theta;x,y)
\right ].
\end{equation}
This formulation preserves the standard full-response OPD objective while providing a natural unit for controlling supervision at the block level.

\vspace{-0.3cm}
\section{Empirical Anatomy of Dense Supervision}
\label{sec:empiral}
\vspace{-0.2cm}

On-policy distillation appears unusually data-efficient partly because a single rollout generates supervision at every autoregressive prefix. A response of length \(N\) does not provide one training target, but a sequence of \(N\) visited states, $s_i := (x,y_{<i}),i=1,\cdots,N$, each paired with a token-level teacher signal. 

Prior work \citep{fu2026oneshotopd} shows that repeated rollouts from even a single query can cover a surprisingly broad region of the state space, and that OPD can continue learning for hundreds of steps even when the training states are fixed. However, \textbf{state count is not necessarily supervision count}. Two visited prefixes may be distinct as states while inducing nearly the same parameter update \citep{shen2026geometryonpolicydistillation,yu2026densesupervisionsparseupdates}; conversely, different tokens may produce distinct gradient directions, but some of these directions may reflect trajectory-specific variation rather than supervision that consistently appears across rollouts \citep{sheng20261tokensenoughgradient}. Thus, the raw supervised-token counts can substantially overestimate the reusable learning signal in a trajectory. We therefore shift from measuring state-space coverage to measuring supervision in gradient space. Rather than asking how many states a rollout visits, we ask a more fundamental question:
\textit{\textbf{How much novel and reproducible learning signal does each additional supervised token actually contribute}}?

\subsection{General Setup}\label{main:sub_steup}

\textbf{Experimental setup.} We study mathematical reasoning and code generation with DeepSeek-R1-Distill-Qwen-1.5B \citep{Guo_DeepSeek-R1_2025} as the student. For mathematics, we use JustRL-DeepSeek-1.5B \citep{he2025justrlscaling15bllm} as the teacher, DAPO-Math-17K \citep{dapo} for training, and AMC 2023 \citep{amc23} and AIME 2025 \citep{aime2025} for evaluation. For code generation, we use DeepCoder-1.5B-Preview \citep{deepcoder2025} as the teacher, TACO \citep{li2023taco} for training, and LiveCodeBench v6 \citep{jain2025livecodebench} (LCB v6) for evaluation. 

\textbf{Hyperparameter settings.} We follow the OPD training pipeline of \citet{fu2026oneshotopd} and hold all optimization hyperparameters fixed across comparisons (details are in Appendix \ref{subapp:hyper}). Specifically, we use rollout and mini-batch sizes of 64 with token-mean loss aggregation. Maximum prompt lengths are 1,024 tokens for mathematics and 4,096 for code, with a maximum response length of 7,168 tokens for both domains. We report avg@4 accuracy on AMC 2023 and AIME 2025, and avg@4 execution accuracy on LCB v6.

\subsection{Fixed-prefix Replay Reveals Rapid Saturation of Supervision Utility}\label{subsec:fixed_prefix}

\textbf{Fixed-prefix replay.}
To investigate how the utility of token-level supervision scales with trajectory length, we follow \citet{fu2026oneshotopd} and construct a controlled fixed-prefix replay setting. \citet{zhao2026needonpolicydistillationdistilling} have also shown that distillation on offline student rollouts can remain effective, providing further support for the use of fixed trajectories in our controlled analysis.
We focus on the medium-difficulty query\footnote{Details of query selection and the scope of the analysis are provided in Appendix \ref{subapp:query}.} used by \citet{fu2026oneshotopd}. This setting provides a non-trivial learning trajectory, where the student improves substantially during training while avoiding early saturation or near-zero success throughout optimization.

For each one-shot query, we sample a rollout batch of $B=64$ trajectories from the initial student policy and keep this trajectory set fixed throughout optimization. Specifically, the replay buffer consists of $B$ sampled trajectories:
\(
\{(x,y^{(b)})\}_{b=1}^{B},
\)
where each response $y^{(b)}$ is sampled independently from $\pi_{\theta_0}(\cdot|x)$.
We then control the amount of accessible supervision by truncating every frozen trajectory at different response prefix lengths,
\(
N\in\{128,512,1024,2048,4096,7168\},
\)
and apply the OPD objective (\eqref{eq:full-token-loss}) only to supervised positions within the retained prefix. These prefix conditions are strictly nested:
\(
\mathcal S_{128}
\subset
\mathcal S_{512}
\subset
\cdots
\subset
\mathcal S_{7168}.
\)
Therefore, increasing $N$ only exposes additional later token-level supervision from the same set of $B$ trajectories, while keeping the query, rollout samples, model initialization, optimization budget, and all other hyperparameters identical. \textbf{Although the replayed states are fixed, the student distribution and teacher--student discrepancy are recomputed under the updated parameters at each optimization step}. As references, we compare against the original on-policy One-Shot OPD, which resamples $B$ trajectories from the evolving student policy, and full-data OPD, which samples $B$ queries from the full training distribution. Unlike these baselines, fixed-prefix replay removes changes in state visitation and isolates the contribution of additional supervised positions along a fixed set of trajectories. This design enables a direct test of whether dense token-level supervision scales proportionally with trajectory length:
\textit{\textbf{
Does exposing more supervised token positions continue to provide proportional learning benefits, or does supervision utility rapidly saturate?
}}

\begin{wrapfigure}{h}{0.58\textwidth}
    \centering
    \vspace{-0.5cm}
    \includegraphics[width=0.28\textwidth]{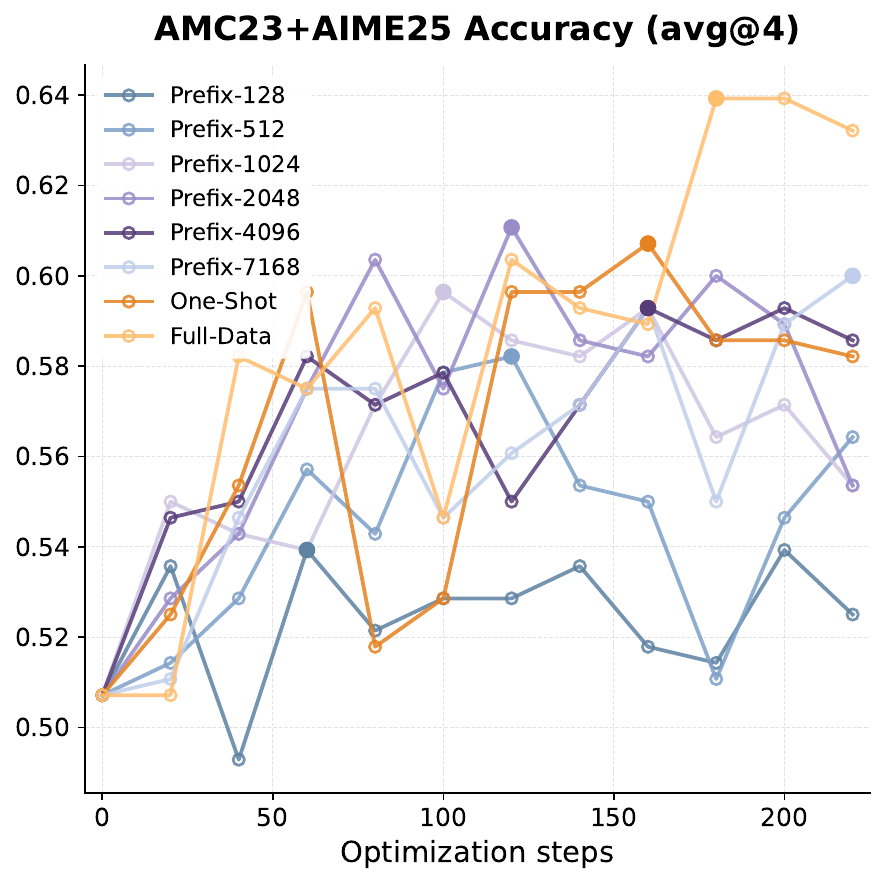}
    \includegraphics[width=0.28\textwidth]{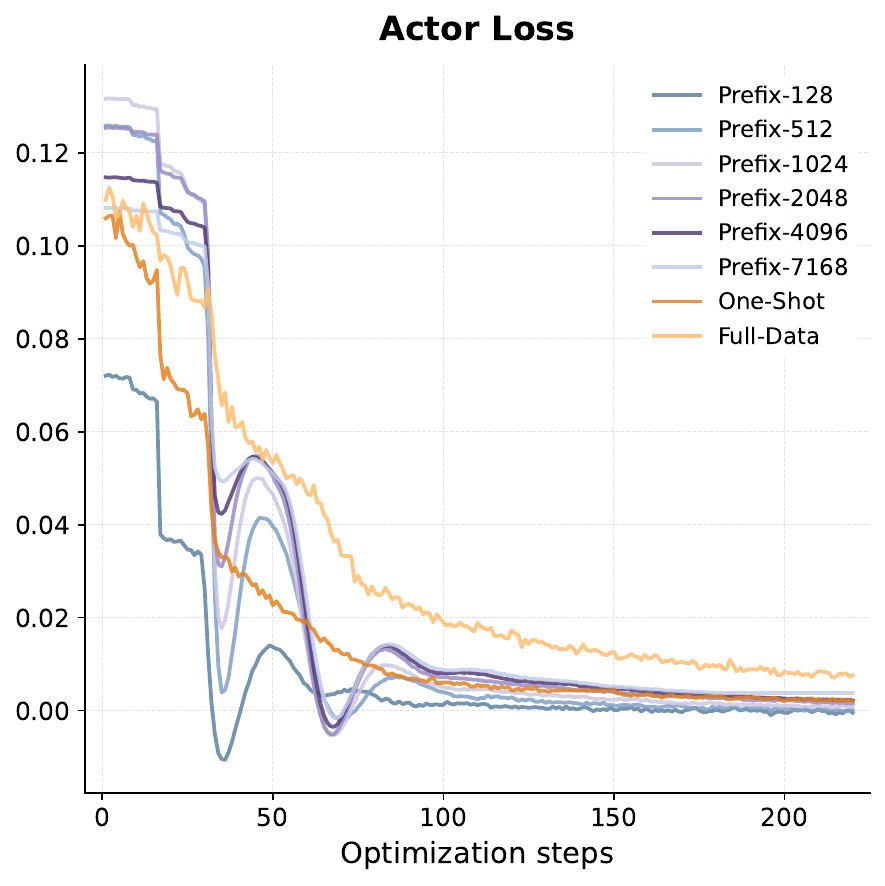}
    \vspace{-0.3cm}
    \caption{Fixed-prefix replay reveals diminishing returns as supervised tokens increase.
\textbf{Left}: validation accuracy averaged over AMC 2023 and AIME 2025 under different prefix lengths.
\textbf{Right}: corresponding actor loss trajectories.
Although longer prefixes expose the student to substantially more supervised token positions, they provide limited additional validation improvement, with later fluctuations reflecting the estimation errors introduced by fixed one-shot replay.}
    \label{fig:fixed_prefix}
    \vspace{-0.45cm}
\end{wrapfigure}

We first investigate whether increasing the amount of token-level supervision along a fixed trajectory leads to proportional improvements in downstream performance. Figure~\ref{fig:fixed_prefix} compares fixed-prefix replay with prefix lengths ranging from 128 to 7168 tokens. As the prefix length increases, the student is exposed to progressively more supervised token positions from the same frozen trajectories. However, the validation performance does not improve proportionally with the number of supervised tokens. In particular, extending the prefix beyond 1024 tokens yields only marginal additional gains on the averaged AMC 2023 and AIME 2025 accuracy, despite introducing substantially more token-level supervision. This effect is not caused by insufficient optimization: all prefix conditions achieve rapid actor loss reduction and similar convergence behavior. Thus, additional supervised tokens yield diminishing downstream returns. These results establish a basic empirical observation:
\(
\text{token count}\uparrow
\not\Rightarrow
\text{validation utility}\uparrow 
\).

\subsection{Position Control Reveals Cross-Block Supervision Transfer}\label{subsec:cross-block}

The saturation observed under fixed-prefix replay admits two non-exclusive explanations: (1) Later trajectory positions may provide weaker or less transferable supervision. 
(2) Supervision from different positions may substantially overlap, so adding later tokens contributes little new information even when those tokens remain individually learnable. 
To disentangle these effects, \textit{we hold the supervised-token budget fixed, vary only the supervised position}, and measure whether optimizing one region of the trajectory also improves teacher--student agreement on other regions.

\textbf{Equal-budget position control.}
We reuse the frozen rollout pool from Section~\ref{subsec:fixed_prefix} for the same query and retain the $B_{\mathrm c}$ trajectories whose responses contain at least 4,096 tokens. Fixing this common trajectory subset across all conditions ensures that positional comparisons are not confounded by differences in trajectory length or sample composition. Within each retained trajectory, we then partition the first 4,096 response tokens into four consecutive, non-overlapping blocks of equal size:
\(
W_k=\mathcal I_k=\{(k-1)m+1,\ldots,km\},
m=1024,\; k\in\{1,2,3,4\}.
\)
Each training condition applies the OPD loss to exactly one block. Tokens preceding the supervised block are retained as causal context but receive no direct supervision. All conditions start from the same initialization $\theta_0$ and are optimized for $T=250$ updates under identical settings. Each update therefore contains exactly $B_{\mathrm c}m$ supervised tokens, ensuring matched supervision and update budgets across conditions.

\textbf{Measuring cross-block transfer.}
Let $\theta_T^{(k)}$ denote the parameters after training on block $W_k$. For each evaluation block $W_j$, we measure the teacher--student discrepancy and its relative reduction from initialization (Positive $R_{k\rightarrow j}$ indicates that training on $W_k$ reduces the discrepancy on $W_j$):
\begin{equation}
\label{eq:block_transfer}
D_j(\theta)
=
\frac{1}{B_{\mathrm c}m}
\sum_{b=1}^{B_{\mathrm c}}
\sum_{i\in\mathcal I_j}
\left|
\log
\frac{\pi_\theta(y_i^{(b)}\mid s_i^{(b)})}
{\pi_{\mathrm t}(y_i^{(b)}\mid s_i^{(b)})}
\right|, \quad
R_{k\rightarrow j}
=
1-\frac{D_j(\theta_T^{(k)})}{D_j(\theta_0)}.
\end{equation}
The discrepancy \(D_j(\theta)\) measures the average absolute difference between the student and teacher log-probabilities, evaluated on the same sampled tokens at fixed autoregressive states in block \(j\). The relative reduction \(R_{k\rightarrow j}\) then quantifies the proportionate decrease in this discrepancy on block \(j\) after optimizing the student with supervision from block \(k\), normalized by its initial value. Accordingly, diagonal entries characterize within-block fitting, whereas off-diagonal entries capture cross-block transfer. Note that this metric quantifies changes in sampled-token log-probability gaps at fixed autoregressive states, rather than measuring the divergence between the full next-token distributions of the student and teacher.

\begin{table}[htb]
    \centering
    \small
\caption{
Cross-block transfer under matched supervision budgets.
Rows denote supervised blocks and columns evaluation blocks; entries report $100R_{k\rightarrow j}$.
Bold diagonal entries show within-block fitting and off-diagonal entries cross-block transfer.
Accuracy is the avg@4 across AMC 2023 and AIME 2025. 
}
    \label{tab:position_transfer}
    \begin{tabular}{lccccc}
        \toprule
        & & \multicolumn{4}{c}{Discrepancy reduction $100R_{k\rightarrow j}$} \\
        \cmidrule(lr){3-6}
        Supervised block
        & Avg@4 Acc. (\%)
        & Eval.\ $W_1$
        & Eval.\ $W_2$
        & Eval.\ $W_3$
        & Eval.\ $W_4$ \\
        \midrule
        $W_1$ & 56.46 & \textbf{92.7} & 63.9 & 53.5 & 46.5 \\
        $W_2$ & 54.38 & 66.8 & \textbf{92.2} & 65.3 & 59.9 \\
        $W_3$ & 55.42 & 60.5 & 70.4 & \textbf{92.0} & 65.8 \\
        $W_4$ & 52.60 & 54.2 & 66.1 & 67.2 & \textbf{91.7} \\
        \bottomrule
    \end{tabular}
\end{table}

(1) \textbf{Later block remains learnable.}
The diagonal entries in Table~\ref{tab:position_transfer} are uniformly high: training on any single block reduces its own teacher--student discrepancy by 91.7\%--92.7\%.
Thus, supervision does not vanish at later trajectory positions; even $W_4$ can be fitted almost as effectively as $W_1$ under the same token and optimization budgets.
Yet downstream accuracy varies across blocks, showing that local learnability does not imply equal validation utility. (2) \textbf{Supervision transfers substantially across positions.}
All off-diagonal entries are positive, with training on one block reducing the discrepancy on unsupervised blocks by 46.5\%--70.4\%.
This transfer extends beyond adjacent windows and operates in both directions, indicating that supervision from different trajectory regions is not independent.
At the same time, within-block reductions remain consistently larger than cross-block reductions, suggesting a combination of cross-position transferable signal and block-specific information.

Taken together, these results rule out a simple ``weak late-token'' explanation for prefix saturation.
\textbf{Later blocks remain highly learnable, yet a substantial fraction of their supervisory signal can already be captured through training elsewhere in the trajectory.}
Thus, adding later tokens increases the number of supervised positions more than it increases the amount of functionally distinct supervision.
While these results establish behavioral transfer across response positions, they do not directly reveal whether such transfer arises from geometrically aligned parameter updates. Nor do they determine whether the remaining position-specific information is reproducible across independent rollouts. We therefore turn to the gradient-space geometry of blockwise OPD supervision.

\subsection{Gradient Geometry Reveals Shared Updates and Heterogeneous Residuals}\label{subsec:geometry}

\begin{figure}[htb]
  \centering
  \includegraphics[width=1.0\textwidth]{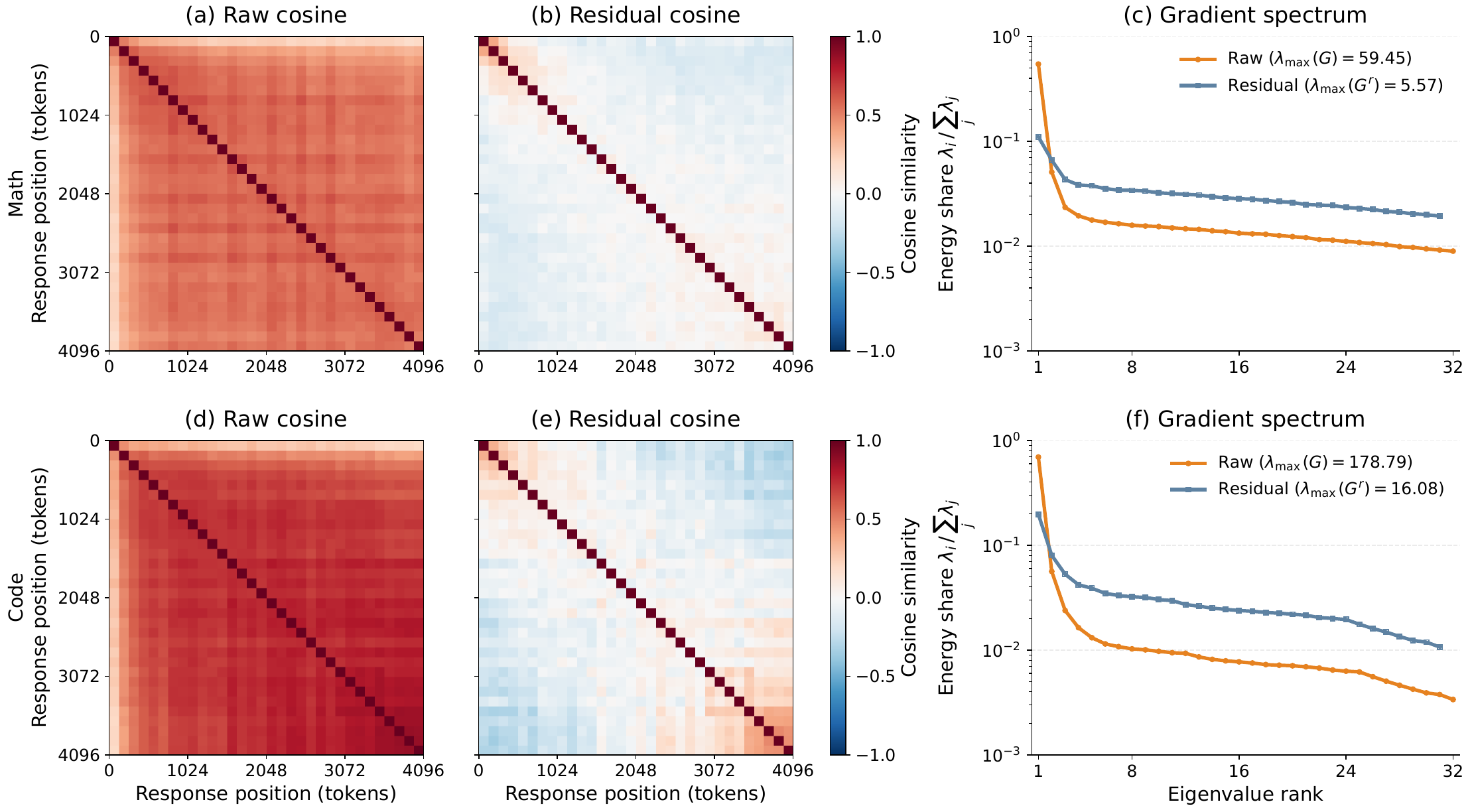}
  \caption{
Blockwise OPD gradient geometry at initialization using 128-token blocks over
the first 4,096 response tokens.
Top: mathematics; bottom: code.
Raw block gradients are positively aligned across all measured positions,
and their spectra contain a dominant component, indicating a global update
direction that persists in multi-prompt batches.
After subtracting the token-weighted mean gradient, residual cosine matrices
exhibit heterogeneous positive and negative structure and substantially flatter
spectra.
Negative residual cosine denotes opposing deviations from the shared mean,
not conflict between the complete block gradients.
The annotated $\lambda_{\max}$ values are the largest eigenvalues of the
unnormalized raw and residual Gram matrices; plotted spectra are normalized
within each matrix.
}
  \label{fig:gradient_geometry}
\end{figure}

The cross-block transfer observed in Section \ref{subsec:cross-block} suggests that supervision from different response positions induces substantially overlapping parameter updates. However, strong transfer does not imply that position-specific supervision is entirely redundant. Different positions may share a dominant update direction while still contributing distinct local corrections. Moreover, the presence of such corrections does not guarantee that they are equally reproducible across independently sampled trajectories. To disentangle these effects, we examine the geometry of blockwise OPD gradients along three dimensions: \textbf{shared directional alignment, position-dependent residual structure, and cross-sample reproducibility}.

\textbf{Blockwise gradient decomposition.} We analyze blockwise OPD gradients at the common initialization \(\theta_0\) for both mathematics and code generation. For each domain, we construct a frozen batch containing one sampled trajectory per distinct prompt, retaining responses with at least 4,096 valid tokens. We partition the first 4,096 tokens into \(K=32\) consecutive blocks of 128 tokens. Let \(g_k\) denote the batch-averaged OPD gradient for block \(k\), computed with detached teacher preference weights and before gradient clipping or optimizer transformations. To distinguish the shared update from position-dependent variation, we decompose each block gradient into its token-weighted mean and a centered residual:

\vspace{-6mm}
\begin{align*}
   \bar g = \sum_{k=1}^{K}w_k g_k,\qquad
\delta_k=g_k-\bar g,\qquad w_k=\frac{1}{K}. 
\end{align*}
\vspace{-5mm}

Since all blocks have equal length, \(\bar g\) coincides with the full-horizon token-mean gradient up to numerical precision. We compare pairwise cosine similarities of the raw gradients and centered residuals, together with the eigenspectra of their respective Gram matrices. Importantly, centering removes the same mean vector from every block; it does not project the gradients onto the orthogonal complement of \(\bar g\).

\ding{182} \textbf{A dominant shared direction persists across positions and prompts}. Figure \ref{fig:gradient_geometry}(a,d) shows substantial positive alignment among raw block gradients throughout the response. The mean pairwise cosine similarity is 0.526 for mathematics and 0.680 for code, while even the minimum similarity remains positive (0.177 and 0.194, respectively). These results indicate that OPD supervision across different response positions contains a strong common update component. Notably, this alignment is observed in batches constructed from distinct prompts, suggesting that the shared direction is not merely an artifact of repeated rollouts of the same question. This provides a gradient-space explanation for the cross-block transfer observed in Sections \ref{subsec:fixed_prefix} and \ref{subsec:cross-block}: optimizing one trajectory region can improve teacher–student agreement elsewhere because \textbf{different regions induce partially aligned parameter updates}.

\ding{183} \textbf{Position-specific supervision remains geometrically diverse}. Subtracting the shared mean reveals a substantially different structure. As shown in Figure \ref{fig:gradient_geometry}(b,e), the residual cosine matrices exhibit heterogeneous positive and negative similarities, with minimum values of -0.174 in mathematics and -0.350 in code. These negative similarities reflect opposing deviations from the common mean, rather than conflicts between the complete block gradients, which remain positively aligned. The residual eigenspectra in Figure \ref{fig:gradient_geometry}(c,f) are also considerably flatter than their raw counterparts. After removing the shared mean, gradient energy is distributed across many directions rather than concentrated in a single dominant mode. Thus, dense supervision cannot be characterized as merely repeated copies of one gradient direction. A substantial amount of position-dependent gradient variation remains beyond the common update. This is consistent with the incomplete cross-block transfer in Section \ref{subsec:cross-block}, where supervision from one block improves other blocks but does not fully replace their direct supervision. \textbf{However, substantial residual variation does not necessarily imply equally reproducible supervision}. While some residual directions may capture systematic, position-specific corrections, others may be highly dependent on individual sampled trajectories. We therefore examine whether these residual directions remain consistent across independent samples.

\ding{184} \textbf{Position-specific residuals exhibit heterogeneous reproducibility}. We split each replay pool into disjoint halves (A) and (B) and independently estimate the blockwise gradients. For each position (k), we measure the alignment of both raw gradients and their deviations from the corresponding half-specific full-horizon gradient: \(\rho_k^{\mathrm{raw}}
=\cos(g_k^A,g_k^B),
\rho_k^{\mathrm{res}}
=\cos(g_k^A-\bar g^A,
g_k^B-\bar g^B).\) For independent rollouts of the same mathematics prompt, the first 128-token residual exhibits strong reproducibility, with \(\rho_1^{\mathrm{res}}=0.787\). This demonstrates that subtracting the shared mean does not eliminate all stable, position-specific structure. In contrast, beyond 1,024 tokens, the mean residual cosine decreases to 0.071, despite a substantial average residual norm of 1.65. Over the same region, raw-gradient cosine remains 0.563. A similar separation appears across disjoint code prompts, where the depth-averaged raw and residual cosine similarities are 0.590 and 0.083, respectively. \textbf{These results show that substantial position-specific gradient variation does not necessarily imply reproducible supervision}. While raw gradients retain strong alignment, residual directions are less consistent across samples, suggesting substantial trajectory-dependent variation. \textbf{Importantly, low residual alignment does not imply that these local corrections are uninformative}.

\textbf{Case study}: Shared reasoning strategies and divergent execution. To contextualize the gradient observations, we inspect $B_c$ independent rollouts of the same mathematical reasoning problem: summing \(1/(2^a3^b5^c)\) over nonnegative integer triples satisfying the even-parity constraint \(a+b+c\equiv0\pmod 2\). Early responses exhibit highly similar structures: all begin with either “Okay” or “Alright,” followed by a restatement of the problem. Importantly, this consistency extends beyond linguistic framing. Within the first 512 tokens, several trajectories mention generating functions, and some explicitly identify the correct parity decomposition. Thus, early alignment is consistent with shared, task-specific reasoning strategies rather than merely superficial formatting.
However, shared strategies do not guarantee consistent execution. One trajectory correctly identifies the parity cases but subsequently reasons about an invalid case with exactly one odd exponent. As generation proceeds, trajectories also diverge in reasoning progress: at the same token depth, some are still exploring solution methods while others have advanced to numerical calculations. Representative excerpts from the sampled trajectories are provided in Table~\ref{tab:qualitative_trajectories}. \textbf{This suggests that weak late-position residual alignment may partly reflect differences in reasoning paths, errors, corrections, and semantic stages, rather than an absence of informative supervision}. Together, these observations highlight a fundamental distinction between the amount of position-specific gradient variation and its reproducibility. \textbf{To scale supervision efficiently, the goal is not to discard later positions, but to reduce redundant reinforcement of shared updates while preserving coverage of potentially useful position-specific corrections across trajectories}.

\section{Stochastic Horizon Supervision: Algorithm and Theory}\label{main:sec_shopd}

The analysis in Section \ref{sec:empiral} reveals two complementary properties of dense OPD supervision. First, \textbf{gradients from different response positions share substantial update directions}, suggesting that supervising every position in every rollout may repeatedly reinforce overlapping information. Second, \textbf{position-specific residuals remain geometrically diverse}, indicating that supervision from earlier positions cannot fully substitute for later supervision. Moreover, the decreasing cross-rollout reproducibility of these residuals suggests that additional depth-specific corrections are not uniformly stable across sampled trajectories, although their low reproducibility does not necessarily imply low utility.

These findings motivate two key design principles: \textbf{reducing redundant reinforcement of shared supervision while preserving access to heterogeneous depth-specific corrections}. Fixed short-horizon supervision can exploit the former but permanently excludes deeper states, whereas full-horizon supervision preserves the latter at substantial computational cost. To balance these competing considerations, we propose \textbf{Stochastic Horizons On-Policy Distillation (SH-OPD)}, which randomly varies the generation horizon across training updates. By frequently sampling shorter horizons while retaining access to longer trajectories, SH-OPD reduces the cost of repeatedly observing overlapping supervision without permanently excluding potentially complementary depth-specific signals. Importantly, SH-OPD adjusts supervision frequency across response depths via stochastic horizon sampling, without explicitly identifying or reweighting residual directions or altering the original token-level OPD objective.

\subsection{Stochastic Horizons On-Policy Distillation}

Prior work has explored horizon control to improve OPD efficiency. POPD progressively expands the rollout horizon, while TOPD adopts a fixed truncated horizon \citep{popd_topd}. More recently, STOPD \citep{stopd} introduces per-rollout stochastic truncation with inverse-probability weighting to compensate for truncation-induced gradient bias. In contrast, SH-OPD adopts a batch-level stochastic horizon shared across rollout workers, avoiding per-sequence horizon sampling and scheduling overhead.

Beyond this implementation difference, SH-OPD takes a different perspective on supervision allocation. Rather than explicitly correcting truncation-induced reweighting to recover the full-horizon update, SH-OPD retains the original token-mean OPD loss and studies the trade-off between reduced computation and the distortion of depth-specific supervision. This choice is motivated by the gradient structure identified in Section \ref{sec:empiral}: substantial cross-position overlap makes uniformly dense supervision potentially redundant, while heterogeneous residuals motivate maintaining access to deeper states. The resulting reweighting bias is characterized in Section \ref{main:subsec_tradeoff}.

\begin{definition}[Power-Survival Horizon Distribution]\label{definition1}
Let $H_{\max}$ denote the maximum response length in standard OPD, and let $H_0$ be a base horizon such that $K=H_{\max}/H_0$ is a positive integer. We define the set of admissible horizons as $\mathcal{H} = \{H_0, 2H_0, \ldots, KH_0\}$. For a parameter $\alpha \geq 0$, the stochastic horizon $H \sim P_\alpha$ follows a discrete power-survival distribution defined by
\begin{equation*}
    \boxed{
    P_\alpha(H \geq kH_0) = k^{-\alpha},
    \quad k=1,\ldots,K.
    }
\end{equation*}
Equivalently, its probability mass function is
\begin{equation*}
    P_\alpha(H=kH_0) =
    \begin{cases}
        k^{-\alpha}-(k+1)^{-\alpha}, & k<K,\\
        K^{-\alpha}, & k=K.
    \end{cases}
\end{equation*}
\end{definition}

\begin{remark}\label{remark1}
The parameter \(\alpha\) controls a trade-off between rollout depth and supervision allocation. At \(\alpha=0\), SH-OPD recovers standard OPD, with \(P_0(H=H_{\max})=1\). For any finite \(\alpha>0\), all admissible horizons have nonzero probability, while increasing \(\alpha\) progressively shifts probability mass toward shorter rollouts, eventually concentrating on \(H_0\) as \(\alpha\to\infty\). For the considered range of \(\alpha\), the horizon probability mass exhibits an approximately U-shaped distribution, favoring shorter and longer horizons over intermediate ones. The expected horizon is
\(
\mathbb E[H_\alpha]=H_0\sum_{k=1}^{K}k^{-\alpha},
\)
yielding an expected nominal generation cost of \(B\mathbb E[H_\alpha]\) per batch, compared with \(BH_{\max}\) under standard OPD\footnote{We discuss this distinction and report empirical termination statistics in Appendix \ref{subapp:horizon_distri}.}. For example, with \(H_0=1024\), \(H_{\max}=7168\), and \(\alpha=0.5\), the expected horizon is approximately \(4{,}114\) tokens, reducing the nominal generation cost by \(42.6\%\). Importantly, stochastic horizon sampling also implicitly reweights supervision across response positions. Assuming all trajectories reach their sampled horizons, the expected weight assigned to block \(j\) is
\(w_{\alpha,j}=\sum_{k=j}^{K}\frac{p_\alpha(k)}{k},\)
where earlier blocks receive higher expected weights due to their inclusion in more sampled horizons. This reduces to uniform weighting, \(w_{0,j}=1/K\), under standard OPD.
\end{remark}

\begin{algorithm}[htb]
\small
\caption{Stochastic Horizons On-Policy Distillation (SH-OPD)}
\label{alg:sh_opd}
\begin{algorithmic}[1]
\Require Student $\pi_\theta$, teacher $\pi_{\mathrm t}$,
dataset $\mathcal D$, batch size $B$,
horizons $(H_0,H_{\max})$, exponent $\alpha$
\State $K\gets H_{\max}/H_0$;
construct horizon distribution $p_\alpha$

\For{each training step $t$}
    \State Sample prompts $\{x_b\}_{b=1}^{B}\sim\mathcal D$

    \Statex \textbf{Batch-level horizon sampling}
    \State Sample \textbf{one shared horizon}:
    $k_t\sim p_\alpha,\quad H_t\gets k_tH_0$

    \Statex \textbf{Parallel rollout generation}
    \State Broadcast $H_t$ to all rollout workers
\For{$b=1,\ldots,B$ \textbf{in parallel}}
    \State $y_b\gets\operatorname{Generate}(\pi_\theta,x_b)$
    \Comment{Terminate at EOS or after $H_t$ tokens}
    \State $N_b\gets |y_b|$
\EndFor

    \Statex \textbf{Standard OPD update}
    \State Compute token losses $\ell_{b,i}(\theta)$
    \State $\displaystyle
    \mathcal L_t\gets
    \frac{\sum_{b=1}^{B}\sum_{i=1}^{N_b}
    \ell_{b,i}(\theta)}
    {\sum_{b=1}^{B}N_b}$
    \State Update $\theta$ using $\nabla_\theta\mathcal L_t$
\EndFor
\end{algorithmic}
\end{algorithm}

\textbf{SH-OPD: Batch-Level Parallel Rollout Generation.} As summarized in Algorithm~\ref{alg:sh_opd}, SH-OPD samples a single horizon per training batch and broadcasts it to all rollout workers. All trajectories are then generated \textbf{in parallel}, stopping at either EOS or the shared horizon $H_t$, rather than generating full-length responses before truncation. The horizon serves only as a generation cap, without conditioning the student policy. All generated tokens are supervised using the standard token-mean OPD loss. This batch-level design avoids per-sequence horizon scheduling overhead while reducing generation cost. Although shorter horizons could enable larger batches under a fixed token budget, we keep the batch size fixed to isolate the effect of stochastic horizon sampling.

\subsection{Theoretical Perspective: A Compute–Reweighting Trade-off}\label{main:subsec_tradeoff}

Section \ref{subsec:geometry} empirically reveals a shared gradient component and position-specific residuals across response blocks. Building on this observation, we extend the decomposition to population-level expected gradients under the current student policy. This allows us to characterize how stochastic horizon sampling reweights supervision across response positions and induces a systematic bias relative to full-horizon OPD.

\begin{assumption}[On-Policy Gradient Decomposition] \label{main:ass1}
At optimization step $t$, SH-OPD samples a horizon
$H_t \sim P_\alpha$ and generates trajectories using the
current student policy $\pi_{\theta_t}$. Let
$\widehat{g}_t^\alpha$ denote the resulting stochastic
gradient of the token-mean OPD surrogate defined in
\eqref{eq:opd-loss}. The student is updated according to
\(\theta_{t+1}
    = \theta_t - \eta \widehat{g}_t^\alpha,
t=1,\ldots,T.\) Let $\mathcal{F}_t$ denote the training history before
sampling the current batch. We define the conditional
expected update direction and its deviation from
full-horizon OPD as
\(    h_\alpha(\theta_t)
    := \mathbb{E}\left[
        \widehat{g}_t^\alpha
        \mid \mathcal{F}_t
    \right], \\
    b_\alpha(\theta_t)
    := h_\alpha(\theta_t) - h_0(\theta_t),\)
where $h_0(\theta_t)$ is the expected full-horizon OPD
update evaluated at the same current parameters
$\theta_t$. Defining
\(\xi_t^\alpha
    = \widehat{g}_t^\alpha - h_\alpha(\theta_t),\)
we obtain
\begin{equation}
\boxed{
\begin{gathered}
    \widehat{g}_t^\alpha
    = h_0(\theta_t)
    + b_\alpha(\theta_t)
    + \xi_t^\alpha, \\
    \mathbb{E}\left[
        \xi_t^\alpha \mid \mathcal{F}_t
    \right] = 0.
\end{gathered}
}
\end{equation}

The decomposition separates the reference OPD update,
the systematic bias induced by stochastic horizons with $\alpha$ in Definition \ref{definition1},
and the stochastic gradient noise. Importantly, all three
components are evaluated at the evolving parameters
$\theta_t$, preserving the on-policy nature of training.    
\end{assumption}

Based on Assumption \ref{main:ass1}, we characterize the horizon-induced bias \(b_\alpha(\theta)\) as follows:

\begin{lemma}[Geometry of the Reweighting Bias]\label{lemma:bias}
For clarity, consider equal-length blocks and trajectories
that reach their sampled horizons. Let
\(\mu_k(\theta)
:=
\mathbb{E}_{\tau \sim \pi_\theta}
\left[
\nabla_\theta \mathcal{L}_k(\theta;\tau)
\right]\)
be the expected gradient of block $k$, with the expectation
taken over current-policy trajectories and the derivative
understood as the detached OPD surrogate gradient. Under this setting,
\(h_0(\theta)
=
\frac{1}{K}\sum_{k=1}^{K}\mu_k(\theta).\) Following Section \ref{subsec:geometry}, define the population-level
centered residual
\(
\delta_k(\theta)
=
\mu_k(\theta)-h_0(\theta).
\)
Let $w_{\alpha,k}$ denote the implicit block weights
described in Remark~\ref{remark1}, and let
$\Delta w_{\alpha,k}=w_{\alpha,k}-1/K$.
Because both weighting schemes sum to one, their shared
gradient component cancels:
\(b_\alpha(\theta)
=
\sum_{k=1}^{K}
\Delta w_{\alpha,k}\mu_k(\theta)
=
\sum_{k=1}^{K}
\Delta w_{\alpha,k}\delta_k(\theta).\) Define the stable residual Gram matrix by
\(\left[G_{\mathrm{stable}}(\theta)\right]_{ij}
=
\left\langle
\delta_i(\theta),\delta_j(\theta)
\right\rangle,\) $\lambda(\cdot)$ denotes the matrix eigenvalue.
The squared systematic bias then admits the exact
characterization:
\begin{equation}
\boxed{
\left\|b_\alpha(\theta)\right\|^2
=
\Delta w_\alpha^\top
G_{\mathrm{stable}}(\theta)
\Delta w_\alpha \leq \lambda_{\text{max}}\left(G_{\mathrm{stable}}(\theta)\right) \|\Delta w_\alpha\|_2^2
}
\label{eq:systematic-bias}
\end{equation}
This result links the gradient geometry in Section \ref{sec:empiral} to the bias induced by stochastic horizon sampling. Since the block weights sum to one, the shared gradient component cancels, and the reweighting bias depends solely on the weighted depth-specific residuals. Unlike the empirical analysis in Section 3.4, which evaluates gradients at initialization, our theoretical characterization is defined at arbitrary current parameters, allowing \(G_{\mathrm{stable}}(\theta)\) to evolve throughout on-policy optimization. Thus, the reweighting bias is jointly determined by the changes in block weights and the geometry of depth-specific residuals, characterizing the trade-off between computational efficiency and preserving depth-specific supervision.
\end{lemma}

The preceding lemma characterizes the systematic
gradient distortion induced by horizon reweighting.
We next examine how this distortion interacts with
stochastic optimization under a finite training budget. 

\begin{lemma}[Stationarity Bound under Horizon Reweighting]
\label{prop:on-policy-bound}

Let $F(\theta)$ be an $L$-smooth, lower-bounded
reference potential satisfying $F(\theta)\ge F_{\inf}$.
We assume that its gradient coincides with the
expected full-horizon OPD update direction, i.e.,
\(
h_0(\theta)=\nabla F(\theta).
\)
Furthermore, assume that the conditional variance
of the stochastic update is uniformly bounded:
\(
\mathbb{E}\!\left[
\|\xi_t^\alpha\|^2\mid\mathcal{F}_t
\right]\le \sigma_\alpha^2.
\)
Under the conditions of Lemma \ref{lemma:bias},
for any constant stepsize $0<\eta\le 1/L$,
\begin{equation}
\begin{aligned}
\frac{1}{T}\sum_{t=1}^{T}
\mathbb{E}\|\nabla F(\theta_t)\|^2
\le {}&
\underbrace{
\frac{2\Delta_F}{\eta T}
}_{\mathrm{optimization}}
+
\underbrace{
L\eta\sigma_\alpha^2
}_{\mathrm{stochastic}}
+
\underbrace{
\mathfrak{B}_{\alpha,T}^2
}_{\mathrm{reweighting}},
\end{aligned}
\label{eq:on-policy-bound}
\end{equation}
where $\Delta_F:=F(\theta_1)-F_{\inf}$ and
\(
\mathfrak{B}_{\alpha,T}^2
:=
\frac{1}{T}\sum_{t=1}^{T}
\mathbb{E}\left[
\Delta w_\alpha^\top
G_{\mathrm{stable}}(\theta_t)
\Delta w_\alpha
\right].
\)
This separates
the optimization error into three terms.
The first decreases with the number of updates,
the second captures stochastic gradient noise,
and the third measures the distortion of stable
depth-specific gradients.
Importantly, the reweighting bias is evaluated
along the evolving optimization trajectory rather
than at a fixed parameter point.
\end{lemma}

\begin{definition}\label{main:definition2}
The benefit of SH-OPD arises from reducing the
expected rollout cost per update.
By Definition \ref{definition1},
\(\mathbb{E}[H_\alpha]
=
H_0\sum_{k=1}^{K}k^{-\alpha}.\)
With a fixed rollout batch size $B$, the expected
nominal generation cost per update is
$c_\alpha=B\mathbb{E}[H_\alpha]$.
Under a total nominal token budget $C$,
the corresponding number of updates is
\(
T_\alpha
=
\left\lfloor
\frac{C}{B\mathbb{E}[H_\alpha]}
\right\rfloor.\)    
\end{definition}

\begin{corollary}[Compute-Normalized Trade-off]
\label{cor:compute-tradeoff}
Under the conditions of
Lemma~\ref{prop:on-policy-bound} and Definition \ref{main:definition2},
assume $\Delta_F>0$ and $C\ge 2c_\alpha$.
Choosing
\(
\eta_\alpha
=
\min\left\{
\frac{1}{L},
\sqrt{
\frac{2\Delta_F}
{L\sigma_\alpha^2T_\alpha}
}
\right\},
\)
where $\eta_\alpha=1/L$ if
$\sigma_\alpha^2=0$, yields

\begin{equation}
\boxed{
\mathcal{E}_\alpha(C)
\lesssim
\underbrace{
\sqrt{
\frac{
\kappa\sigma_\alpha^2
\mathbb{E}[H_\alpha]
}{C}
}
}_{\text{finite-compute stochastic cost}}
+
\underbrace{
\mathfrak{B}_{\alpha,T_\alpha}^2
}_{\text{reweighting bias}},
}
\label{eq:compute-bias-tradeoff}
\end{equation}
where $\mathcal{E}_\alpha(C)
:=
\frac{1}{T_\alpha}
\sum_{t=1}^{T_\alpha}
\mathbb{E}\|\nabla F(\theta_t)\|^2$, and $\kappa=16L\Delta_F B$ is independent
of $\alpha$ and $C$.
\end{corollary}

\textbf{How to Choose $\alpha$?} Corollary \ref{cor:compute-tradeoff} establishes a compute--reweighting trade-off: shorter expected horizons permit more on-policy updates under a fixed compute budget, potentially reducing stochastic optimization error, but may also distort reproducible depth-specific gradient structure, as characterized by Lemma \ref{lemma:bias}. While $\alpha=0$ recovers full-horizon OPD with zero reweighting bias ($\mathfrak{B}_{\alpha,T_\alpha}^2=0$), increasing $\alpha$ is beneficial when the reduction in compute-adjusted stochastic cost outweighs the additional bias (Appendix \ref{app:sub_compare}). Thus, $\alpha$ governs a principled balance between computational efficiency and the preservation of stable supervision structure. In the next part, we investigate the choice of \(\alpha\) empirically.

\section{Empirical Evaluation}\label{sec:sh_ablation}

\textbf{Experimental setup.} We evaluate SH-OPD on mathematical reasoning and code generation with \(H_0=1024\) and \(H_{\max}=7168\). We consider \(\alpha\in\{0,0.5,1.0,1.5,5.0\}\) for both mathematics and code generation. All other experimental settings follow Section \ref{main:sub_steup}, with the horizon distribution being the only factor varied within each task. Notably, \(\alpha=0\) exactly recovers standard full-horizon OPD. Additional results extending SH-OPD to multi-teacher on-policy distillation (MOPD) and Instruction following tasks are presented in the Appendix \ref{app:sub_mopd}.

\begin{figure*}[t]
  \centering
  \includegraphics[width=0.98\textwidth]{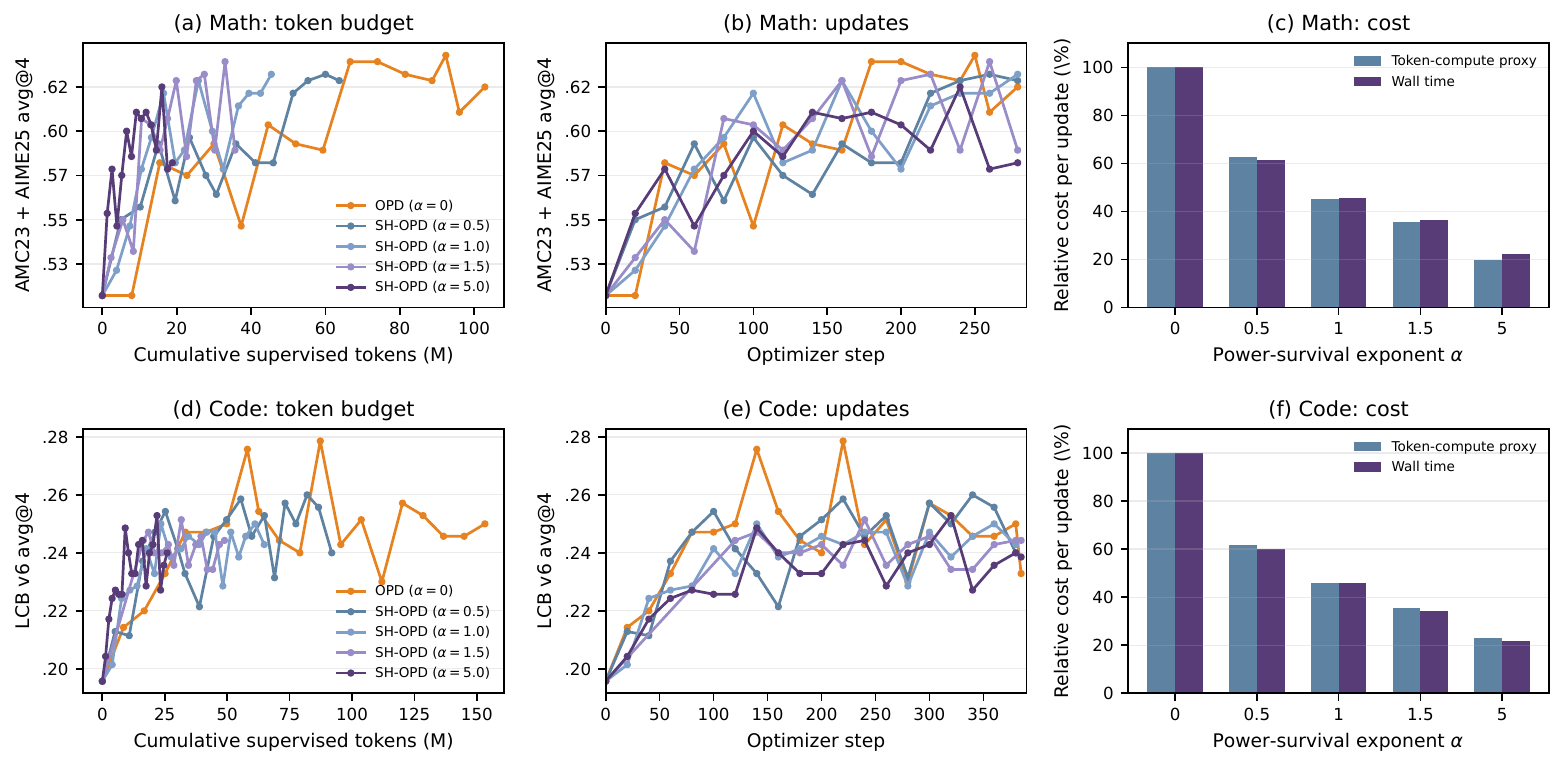}
  \caption{
  \textbf{Performance–cost trade-off under stochastic horizons.}
  Mathematics (top) and code generation (bottom) are evaluated across different horizon exponents \(\alpha\). The left column shows validation accuracy against cumulative supervised response tokens, the middle column shows accuracy against optimization steps, and the right column reports per-update token-compute proxy and wall-time cost relative to standard OPD (\(\alpha=0\)). SH-OPD substantially reduces training cost while maintaining competitive performance. Mathematics favors \(\alpha=1.0\), whereas code generation prefers \(\alpha=0.5\), highlighting the task-dependent trade-off between supervision efficiency and depth coverage.
  }
  \label{fig:sh_opd_main}
\end{figure*}

\textbf{Efficiency metrics.}
We measure training efficiency using two complementary per-update costs. The \textit{token-compute proxy} counts the total prompt and response tokens processed by the rollout and training pipeline per update. It approximates sequence-processing workload but does not account for attention complexity or hardware utilization. The \textit{wall-time} cost measures the end-to-end duration of each training update, including rollout generation, student and teacher scoring, and actor optimization, but excluding validation. Both metrics are averaged over logged training updates and normalized to standard OPD within each domain. In contrast, the \textit{cumulative supervised tokens} in Figure \ref{fig:sh_opd_main}(a,d) count only valid generated response tokens contributing to the OPD loss, accounting for natural EOS termination.

\textbf{SH-OPD improves supervision efficiency.}
Figure~\ref{fig:sh_opd_main} demonstrates that stochastic horizons substantially reduce training cost while preserving competitive downstream performance. On mathematics, \(\alpha=1.0\) achieves the strongest overall trade-off. \textit{At approximately 45 million supervised response tokens, it reaches 63.21\% accuracy, compared with 60.36\% for standard OPD at a similar token budget (44.6 million tokens)}. By the end of training, it also achieves a higher final accuracy than standard OPD (63.21\% vs. 62.50\%), while reducing the per-update token-compute proxy to 45.1\% and wall time to 45.5\% of the standard OPD baseline. These results demonstrate that reducing redundant supervision can improve learning efficiency without sacrificing solution quality.

\textbf{The preferred exponent differs across tasks.}
Code generation favors a more conservative horizon distribution. Among the tested SH-OPD settings, \(\alpha=0.5\) achieves a higher peak accuracy than \(\alpha=1.0\) (26.00\% vs. 25.00\%) and a higher mean accuracy over the final five evaluations (25.26\% vs. 24.49\%). Although their final checkpoints are close, the overall learning trajectory favors \(\alpha=0.5\). This setting reduces the per-update token-compute proxy to 61.8\% and wall time to 60.0\% of standard OPD, demonstrating substantial efficiency gains while maintaining competitive code-generation performance.

\textbf{More aggressive horizon reduction is not always beneficial.} The mathematics sweep further reveals that increasing \(\alpha\) beyond 1.0 does not consistently improve performance. Although \(\alpha=1.5\) and \(\alpha=5.0\) further reduce the cost per update, their final accuracies decrease to 58.93\% and 58.21\%, respectively. Thus, shorter horizons do not necessarily provide better supervision efficiency: excessively concentrating supervision on early response positions can compromise learning quality.

\textbf{Choosing $\alpha$ through Gradient Geometry.} The task-dependent preference for horizon exponents is consistent with the compute–reweighting trade-off in Section \ref{main:subsec_tradeoff}. By exploiting shared gradients, SH-OPD reduces redundant supervision, while aggressive horizon reduction may distort complementary depth-specific updates. Lemma \ref{lemma:bias} shows that this distortion is governed by the Gram matrix of expected residual gradients. Empirically, mathematics favors $\alpha=1.0$, whereas code generation prefers $\alpha=0.5$, retaining more frequent deep supervision. This difference is qualitatively consistent with the larger residual Gram eigenvalue observed for code (16.08) than for mathematics (5.57), suggesting greater sensitivity to depth reweighting. However, these eigenvalues are measured from finite-batch gradients over the first 4,096 tokens at initialization and do not directly quantify the population-level bias over the full training horizon. Thus, the observed gradient geometry provides a plausible explanation for the task-dependent trade-off rather than a quantitative prediction of the optimal exponent.

Overall, the results demonstrate that effective OPD supervision does not require uniformly dense coverage of every response position. SH-OPD achieves substantial computational savings by reducing the frequency of full-horizon supervision while retaining access to deeper states. The different preferences for horizon reduction across tasks further highlight the importance of balancing computational efficiency with the preservation of complementary depth-specific supervision.

\section{Related Work}

\textbf{OPD and Learning Dynamics.} On-policy distillation (OPD) transfers teacher capabilities through token-level supervision on student-generated trajectories, mitigating the distribution mismatch of off-policy distillation \citep{agarwal2024onpolicy,gu2024minillm}. Recent studies have examined the factors governing its effectiveness. \citet{li2026rethinking} identify teacher--student compatibility and novel teacher knowledge as important conditions for successful distillation. \citet{ding2026doesonpolicydistillationreally} question the role of teacher knowledge transfer in OPD, showing that substantial performance gains can persist under noisy teacher supervision and can be partially reproduced through teacher-free self-improvement. \citet{fu2026oneshotopd} show that repeated rollouts from a single query can cover a substantial fraction of the states visited during full-data training, while the absorption of teacher supervision remains slow. Other works investigate the reliability of the distillation objective. Entropy-Aware OPD and Hybrid Policy Distillation adjust divergence objectives to improve distribution matching and training stability \citep{jin2026entropy,zhu2026hybrid}. \citet{fu2026revisiting} analyze the bias and variance of sampled-token supervision and identify failure modes arising from long-horizon prefix drift. Similarly, PowerOPD addresses unstable token-level learning signals through bounded reward transformations \citep{zhao2026poweropd}. These studies characterize the conditions and optimization mechanisms underlying OPD. In contrast, our work focuses on how supervision from different trajectory positions interacts and contributes to learning.

\textbf{Supervision Allocation in OPD.} Recent studies have explored selective supervision to improve OPD efficiency. TIP identifies informative tokens using student entropy and teacher--student divergence \citep{xu2026tiptokenimportanceonpolicy}, while teachability-aware selection and information-efficiency criteria further account for local learnability and gradient-estimation reliability \citep{wang2026teachability,sheng2026ier}. Position-aware methods also adjust token weights according to accumulated teacher--student discrepancies \citep{xie2026positionbiasonpolicydistillation}. These approaches primarily optimize which generated tokens receive supervision. A complementary line of work reduces rollout generation costs through horizon control. FastOPD demonstrates that reasoning-prefix distillation can match full-trajectory OPD with substantially lower computation \citep{zhang-etal-2026-fast}. POPD and TOPD adopt progressive and fixed rollout truncation, respectively \citep{popd_topd}, while temporal curriculum methods progressively extend trajectory depth for multi-turn agents \citep{wang2026tcod}. More recent approaches adapt the horizon to training conditions: KAT terminates rollouts exhibiting persistent low-KL agreement \citep{xin-etal-2026-escaping}, while ADWIN selects prefix windows using delayed full-rollout probes and prefix--full gradient alignment \citep{liang-etal-2026-adwin}. TurnOPD further introduces adaptive rollout-depth and turn-balanced supervision for long-horizon agents \citep{zhou2026turnopd}.

\textbf{Gradient Geometry and Supervision Efficiency.} Recent studies have investigated the geometry of OPD updates, revealing subspace locking \citep{shen2026geometryonpolicydistillation}, gradient sparsity and spectral concentration \citep{yu2026densesupervisionsparseupdates}, and token-level gradient-estimation reliability \citep{sheng2026ier}. ADWIN further exploits prefix--full gradient alignment for adaptive horizon selection \citep{liang-etal-2026-adwin}. Beyond these analyses, we investigate cross-position gradient structure, identifying shared update components, heterogeneous position-specific residuals, and their consistency across sampled trajectories. We further characterize horizon-induced reweighting bias through the Gram matrix of expected depth-specific gradient residuals, establishing a theoretical connection between gradient geometry and the trade-off between compute-adjusted optimization error and supervision reweighting bias. \textit{While prior work has improved OPD efficiency through selective supervision and horizon control, our work connects cross-position supervision transfer and gradient residual structure to the bias–compute trade-off of stochastic horizon allocation}.

\section{Conclusion} 
{\small
\[
\boxed{
\begin{gathered}
\text{Dense supervision exhibits diminishing returns}\\
\downarrow\\
\text{Shared gradients + heterogeneous residuals}\\
\downarrow\\
\text{Sparse deep supervision with stochastic horizons}\\
\downarrow\\
\text{Compute--reweighting trade-off}
\end{gathered}
}
\]
}

In this work, we revisit the scaling of token-level supervision in on-policy distillation. Through controlled experiments and gradient-space analysis, we show that increasing supervision density yields diminishing returns, as gradients across response positions share substantial update directions while retaining heterogeneous, less reproducible depth-specific components. \textbf{Dense supervision should be reduced where supervision overlaps, while complementary depth-specific supervision should remain accessible rather than be permanently truncated}. Motivated by these findings, we introduce \textbf{Stochastic Horizons On-Policy Distillation (SH-OPD)}, which reduces redundant supervision by randomly varying rollout horizons while preserving access to deeper response states. Our theoretical analysis further characterizes the trade-off between compute-adjusted stochastic optimization cost and the distortion of stable depth-specific gradient structure. Experiments on mathematical reasoning and code generation demonstrate that SH-OPD substantially reduces training cost while maintaining competitive or improved downstream performance. The preferred horizon exponent differs across tasks, consistent with their heterogeneous gradient structures. These findings suggest that \textbf{effective supervision scaling is not simply about increasing the number of supervised tokens, but about allocating supervision efficiently across response depths}. More broadly, SH-OPD offers a simple approach to improving the compute efficiency of on-policy distillation without requiring uniformly dense supervision.

\newpage

\bibliography{main}


\appendix
\onecolumn

\section{Extension to Experiments}

\subsection{Default hyperparameters for OPD}\label{subapp:hyper}

\textbf{Training and evaluation.} We follow the default OPD configuration of \citet{fu2026oneshotopd}, using AdamW with a learning rate of $1\times10^{-6}$, weight decay 0.01, a rollout and mini-batch size of 64, sampling temperature 1.0, top-p=1.0, token-mean loss aggregation and gradient clipping at 1.0. We use maximum prompt lengths of 1,024 and 4,096 tokens for mathematics and code, respectively, and a maximum response length of 7,168 tokens for both domains. We evaluate each problem with four sampled responses, reporting avg@4 accuracy on AMC 2023 and AIME 2025 and avg@4 execution accuracy on LiveCodeBench v6. All comparisons use identical optimization hyperparameters and evaluation budgets.  All experiments are conducted on 4 NVIDIA H20 GPUs.

\subsection{Medium query}\label{subapp:query}
\textbf{Query selection and difficulty.} We adopt the medium-difficulty mathematical reasoning query (ID: t22) from the One-Shot OPD study by \citet{fu2026oneshotopd}. The query is drawn from DAPO-Math-17K \citep{dapo}, where its difficulty is determined by the initial student's pass rate over eight independently sampled responses before training. Specifically, the medium query is solved in 4 out of 8 attempts, corresponding to an initial pass rate of 50\%. The original study further shows that this query exhibits a gradual improvement in training success rate, from approximately 0.36 to 0.89 over 300 optimization steps, whereas the easy query is consistently solved and the hard query remains unsolved. We select the medium query because it provides a non-trivial learning regime with sufficient room for improvement, allowing us to examine how supervision from different response positions contributes to downstream learning without being dominated by early saturation or persistent failure.

\textbf{Full query}. The selected query is: Let \(S\) be the set of triples \((a,b,c)\) of non-negative integers such that \(a+b+c\) is even. Determine the value of

\[
\sum_{(a,b,c)\in S}\frac{1}{2^a3^b5^c}.
\]

This sum can be expressed as \(m/n\), where \(m\) and \(n\) are relatively prime positive integers. Compute \(m+n\).

\textbf{Fixed-prefix replay construction.} Following the fixed-state analysis of \citet{fu2026oneshotopd}, we construct a frozen replay pool by sampling \(B=64\) independent trajectories from the initial student policy on this query, using a sampling temperature of 1.0 and a maximum response length of 7,168 tokens. The sampled trajectories are stored and reused throughout optimization, so the visited autoregressive states remain fixed while the student parameters evolve. For the fixed-prefix experiment in Section \ref{subsec:fixed_prefix}, we consider supervision horizons of \(\{128,512,1024,2048,4096,7168\}\) tokens. Each condition uses the same replay pool but restricts the OPD loss to valid token positions within the corresponding prefix. Trajectories shorter than a specified prefix contribute only their available valid tokens. Thus, the supervision sets are nested, and increasing the prefix length introduces additional later-position supervision without changing the underlying trajectories. At each optimization step, the teacher–student log-probability discrepancy is evaluated on the retained states using the updated student parameters.

\textbf{Common trajectory subset for position-controlled experiments}. For the cross-block transfer analysis in Section \ref{subsec:cross-block}, we further restrict the replay pool to trajectories containing at least 4,096 valid response tokens. Denoting the number of retained trajectories by \(B_c\), we partition the first 4,096 tokens of each retained trajectory into four non-overlapping blocks of 1,024 tokens. All position-controlled conditions share this identical trajectory subset, ensuring that differences in training outcomes arise from the supervised positions rather than trajectory length or sample composition. Each condition optimizes one block for 250 updates, with preceding tokens preserved as causal context but excluded from direct supervision.

\textbf{Scope of the analysis.} \citet{fu2026oneshotopd} show that the one-shot OPD phenomenon is robust to variations in query difficulty, response-length cap, and sampling temperature. \textbf{Our experiments build on this controlled setting but address a different question: rather than examining how many distinct states are sufficient for learning, we investigate how much additional supervision is contributed by different token positions within a fixed set of trajectories}.

\subsection{Qualitative trajectory excerpts}

\begin{table}[htb]
\centering
\footnotesize
\setlength{\tabcolsep}{4pt}
\caption{
Qualitative excerpts from independent rollouts of the same medium-difficulty query from the initial student. Trajectory IDs are zero-based row indices in the frozen pool, and intervals are half-open response-token ranges. Excerpts are verbatim apart from ellipses and light mathematical typesetting. 
}
\label{tab:qualitative_trajectories}
\begin{tabular}{L{0.08\linewidth}L{0.13\linewidth}L{0.49\linewidth}L{0.20\linewidth}}
\toprule
Stage & Sampled trajectory & Excerpt & Qualitative observation \\
\midrule
Early
& ID 1, $[128,512)$
& ``The problem reminds me of generating functions where you can encode conditions on the exponents \ldots{} Because the condition is \(a+b+c\) is even, it splits into two cases: either all of \(a,b,c\) are even, or exactly two of them are odd.''
& Shared task-specific strategy and correct parity decomposition. \\
\addlinespace
Early
& ID 4, $[256,448)$
& ``Case 2: Exactly two of \(a,b,c\) are odd \ldots{} For Case 2: For the case where exactly one is odd, we have \(a=2a'+1\), \(b=2b'\), \(c=2c'\) or variations.''
& A correct decomposition immediately branches into an invalid one-odd case. \\
\addlinespace
Middle
& ID 20, $[1792,1984)$
& ``Hmmm. Hmmm confusing. Wait no, maybe I should model \(U\) separately \ldots{} How can we relate \(U\) to \(T\)? \ldots{} Wait, hold on \ldots{} Wait, no.''
& Still reorganizing the parity argument. \\
\addlinespace
Middle
& ID 41, $[1792,2048)$
& ``\(25/12+5/3=45/12=15/4\), which is correct \ldots{} Thus, the answer is \(25/12\). Therefore, \(m=25\), \(n=12\), so \(m+n=37\).''
& Already completes the numerical consistency check at a comparable depth. \\
\addlinespace
Late
& ID 12, $[3328,3584)$
& ``Recall that \(G(q)=\) even part \(+\) odd part \ldots{} probably get \(25/12\) \ldots{} \(m+n\) is \(25+12=37\).''
& Reaches the correct final result after extended verification. \\
\addlinespace
Late
& ID 44, $[3200,3456)$
& ``\(T_{\mathrm{even}}=31/16\), \(T_{\mathrm{odd}}=29/16\) \ldots{} Thus, the sum is \(31/16\) \ldots{} \(m+n=47\).''
& Reaches a self-consistent but incorrect branch and final answer. \\
\bottomrule
\end{tabular}
\end{table}

Table~\ref{tab:qualitative_trajectories} complements the case study in Section~\ref{subsec:geometry} with excerpts from the frozen rollout pool. The examples illustrate three recurring patterns: a shared generating-function strategy early in generation, trajectory-specific mistakes despite initially correct parity reasoning, and different levels of progress at comparable response depths.

\subsection{Horizon Distribution Statistics}\label{subapp:horizon_distri}
\textbf{Power-survival horizon distribution.} We use \(H_0=1024\) and \(H_{\max}=7168\), yielding seven admissible horizons. Under the power-survival distribution, the expected horizons for \(\alpha\in\{0,0.5,1.0,1.5,5.0\}\) are 7,168, 4,114, 2,655, 1,928, and 1,062 tokens, respectively, corresponding to nominal generation-budget reductions of 0\%, 42.6\%, 63.0\%, 73.1\%, and 85.2\% relative to full-horizon OPD. The probability of sampling the maximum horizon decreases from 100\% at \(\alpha=0\) to 37.8\%, 14.3\%, 5.4\%, and 0.006\%, respectively. Thus, increasing \(\alpha\) progressively concentrates sampling on shorter horizons while retaining nonzero probability of deeper supervision. These statistics characterize nominal horizon budgets; actual generated lengths may be shorter due to EOS termination.

\begin{table}[htb]
\centering
\footnotesize
\setlength{\tabcolsep}{4pt}
\caption{Statistics of the power-survival horizon distribution
with $H_0=1024$ and $H_{\max}=7168$.
Token reduction is computed relative to full-horizon OPD
($\alpha=0$), assuming trajectories reach their sampled horizons.}
\label{tab:horizon_statistics}
\begin{tabular}{ccccc}
\toprule
$\alpha$ & $\mathbb{E}[H]$ & Token Reduction
& $P(H=H_0)$ & $P(H=H_{\max})$ \\
\midrule
0   & 7,168 & 0.0\%  & 0.00\%  & 100.00\% \\
0.5 & 4,114 & 42.6\% & 29.29\% & 37.80\% \\
1.0 & 2,655 & 63.0\% & 50.00\% & 14.29\% \\
1.5 & 1,928 & 73.1\% & 64.64\% & 5.40\% \\
5.0 & 1,062 & 85.2\% & 96.88\% & 0.006\% \\
\bottomrule
\end{tabular}
\end{table}

As shown in Table \ref{tab:horizon_statistics}, increasing \(\alpha\) progressively shifts probability mass toward shorter horizons, substantially reducing the expected generation budget. Notably, moderate values of \(\alpha\) retain meaningful probabilities of sampling the maximum horizon, whereas \(\alpha=5.0\) concentrates 96.88\% of the probability mass on the shortest horizon, effectively approaching fixed short-horizon supervision.

\textbf{Empirical sampled and realized response lengths.}
Table~\ref{tab:empirical_horizon_statistics} reports statistics from the full-data training runs used in Figure~\ref{fig:sh_opd_main}. A single horizon is sampled per global update and shared by all \(B=64\) trajectories in that rollout batch. We define the empirical sampled horizon as the arithmetic mean of these per-update samples, and the realized response length as the mean number of valid generated response tokens after natural termination. The empirical sampled horizons are close to their theoretical expectations, with finite-run deviations caused by sampling variability. Realized response lengths are lower than the sampled horizons because some trajectories terminate at EOS. 

\begin{table}[htb]
\centering
\scriptsize
\setlength{\tabcolsep}{2pt}
\renewcommand{\arraystretch}{0.9}
\caption{
Empirical horizon and response-length statistics for the full-data runs.
\(\mathbb{E}[H]\) is the theoretical mean sampled horizon,
\(\overline{H}_{\mathrm{sampled}}\) is its empirical mean across logged updates,
and \(\overline{L}_{\mathrm{response}}\) is the mean realized response length after EOS termination.
}
\label{tab:empirical_horizon_statistics}
\begin{tabular}{lrrrrrr}
\toprule
Domain & \(\alpha\) & Updates & \(\mathbb{E}[H]\)
& \(\overline{H}_{\mathrm{sampled}}\)
& \(\overline{L}_{\mathrm{response}}\) & EOS term. \\
\midrule
Math & 0   & 279 & 7,168 & 7,168 & 5,764 & 54.34\% \\
Math & 0.5 & 279 & 4,114 & 4,169 & 3,569 & 25.25\% \\
Math & 1.0 & 279 & 2,655 & 2,859 & 2,546 & 13.88\% \\
Math & 1.5 & 279 & 1,928 & 2,184 & 1,999 & 8.70\% \\
Math & 5.0 & 279 & 1,062 & 1,057 & 1,055 & 0.58\% \\
\midrule
Code & 0   & 384 & 7,168 & 7,168 & 6,452 & 23.67\% \\
Code & 0.5 & 384 & 4,114 & 4,091 & 3,783 & 11.58\% \\
Code & 1.0 & 384 & 2,655 & 2,811 & 2,655 & 6.67\% \\
\bottomrule
\end{tabular}
\end{table}

\subsection{Multi-Teacher On-Policy Distillation}\label{app:sub_mopd}

\textbf{Setup.}
We extend SH-OPD to multi-teacher on-policy distillation (MOPD) with a single DeepSeek-R1-Distill-Qwen-1.5B student. Each example is routed by its ability label to a frozen domain-specific teacher: JustRL-DeepSeek-1.5B for 17,917 DAPO-Math training examples, DeepCoder-1.5B-Preview for 24,701 TACO code examples, and UltraData-IF-1.5B \citep{fu2026oneshotopd} for 4,501 first-turn Multi-IF instruction-following examples \citep{ultradata-sft-2605}. We compare \(\alpha\in\{0,0.5,1.5\}\) using \(H_0=1024\) and \(H_{\max}=7168\). Mathematics and code are evaluated during training with avg@1 on AMC 2023 (40 problems), AIME 2025 (30 problems), and LiveCodeBench v6 (175 problems), using a 31,744-token response cap. Instruction following is evaluated on all 4,445 Multi-IF conversations, each containing three turns in one of eight languages.

\begin{table}[htb]
\centering
\scriptsize
\setlength{\tabcolsep}{3pt}
\caption{
MOPD efficiency and downstream performance. Per-update token-compute and wall-time costs are averaged over updates 1--300 and normalized to full-horizon MOPD. Math and code results are the mean avg@1 over the last three checkpoints, while Multi-IF scores are reported at checkpoint 300.
}
\label{tab:mopd_efficiency}
\begin{tabular}{crrrrrrr}
\toprule
\(\alpha\)
& Token cost
& Wall time
& AMC23
& AIME25
& LCB v6
& Multi-IF mean
& Multi-IF turn 3 \\
\midrule
0   & 100.0\% & 100.0\% & \textbf{85.83} & \textbf{36.67} & \textbf{25.33} & 37.89 & 26.79 \\
0.5 & 61.2\%  & 59.9\%  & 83.00 & 32.22 & \textbf{25.33} & \textbf{39.26} & \textbf{27.81} \\
1.5 & 36.4\%  & 35.6\%  & 82.50 & 35.56 & 23.05 & 38.73 & 26.96 \\
\bottomrule
\end{tabular}
\end{table}

\textbf{Results.}
SH-OPD remains effective under multi-teacher distillation. With $\alpha=0.5$, SH-OPD reduces per-update token-compute and wall-time costs to 61.2\% and 59.9\% of full-horizon MOPD, respectively. It matches the full-horizon result on LCB v6 and improves the Multi-IF mean from 37.89 to 39.26, while incurring moderate decreases on the two mathematics benchmarks. The more aggressive $\alpha=1.5$ setting reduces both costs by roughly 64\%, but exhibits less consistent performance across tasks. Overall, stochastic horizons retain useful multi-domain performance at substantially lower training cost, extending the efficiency benefits of SH-OPD beyond the single-teacher setting. Since Multi-IF training and evaluation share source conversations, the Multi-IF results should be interpreted as in-sample performance comparisons rather than evidence of held-out generalization.

\newpage

\section{Omitted Proofs and Additional Results}

\subsection{Proof of Lemma 1}
\label{app:compute-reweighting}

\begin{proof}
We derive the population-level reweighting identity
and establish the corresponding quadratic-form
characterization of the systematic bias.

Consider $K$ equal-length blocks, each containing
$H_0$ tokens. Let
\[
p_\alpha(m)
=
\mathbb{P}_\alpha(H=mH_0),
\qquad m=1,\ldots,K,
\]
denote the probability of sampling a horizon
comprising $m$ blocks. As assumed in the lemma,
trajectories reach their sampled horizons, and
horizon sampling preserves the distribution of
the observed prefixes. Consequently, the expected
gradient of block $k$, whenever included, is
$\mu_k(\theta)$.

For a sampled horizon $H=mH_0$, the token-mean
gradient is the average of the first $m$ block
gradients. Taking the expectation over sampled
horizons yields
\begin{align}
h_\alpha(\theta)
&=
\sum_{m=1}^{K}
p_\alpha(m)
\left(
\frac{1}{m}
\sum_{k=1}^{m}\mu_k(\theta)
\right)
\nonumber\\
&=
\sum_{k=1}^{K}
\underbrace{
\left(
\sum_{m=k}^{K}
\frac{p_\alpha(m)}{m}
\right)
}_{w_{\alpha,k}}
\mu_k(\theta).
\label{eq:app-horizon-weights}
\end{align}
Here, $w_{\alpha,k}$ represents the effective
weight assigned to block $k$ under the sampled
horizon distribution. These weights are normalized,
since
\begin{align}
\sum_{k=1}^{K}w_{\alpha,k}
&=
\sum_{m=1}^{K}
\frac{p_\alpha(m)}{m}
\sum_{k=1}^{m}1
\nonumber\\
&=
\sum_{m=1}^{K}p_\alpha(m)
=1.
\end{align}

When $\alpha=0$, the horizon is fixed at
$H_{\max}=KH_0$. Thus, $w_{0,k}=1/K$ for all $k$,
and
\[
h_0(\theta)
=
\frac{1}{K}
\sum_{k=1}^{K}\mu_k(\theta).
\]

Now define the weight perturbation
$\Delta w_{\alpha,k}=w_{\alpha,k}-1/K$.
By normalization,
$\sum_{k=1}^{K}\Delta w_{\alpha,k}=0$.
Recalling the centered population residual
$\delta_k(\theta)=\mu_k(\theta)-h_0(\theta)$,
we obtain
\begin{align}
b_\alpha(\theta)
&=
h_\alpha(\theta)-h_0(\theta)
\nonumber\\
&=
\sum_{k=1}^{K}
\Delta w_{\alpha,k}\mu_k(\theta)
\nonumber\\
&=
\sum_{k=1}^{K}
\Delta w_{\alpha,k}
\bigl(
h_0(\theta)+\delta_k(\theta)
\bigr)
\nonumber\\
&=
\sum_{k=1}^{K}
\Delta w_{\alpha,k}\delta_k(\theta).
\label{eq:app-bias-decomposition}
\end{align}
The last equality follows from the zero-sum
property of the weight perturbations, which
eliminates the shared gradient component.

By the definition of the stable residual Gram
matrix,
\[
[G_{\mathrm{stable}}(\theta)]_{ij}
=
\langle
\delta_i(\theta),\delta_j(\theta)
\rangle,
\]
the squared bias can be expressed as
\begin{align}
\|b_\alpha(\theta)\|^2
&=
\left\|
\sum_{k=1}^{K}
\Delta w_{\alpha,k}\delta_k(\theta)
\right\|^2
\nonumber\\
&=
\sum_{i=1}^{K}\sum_{j=1}^{K}
\Delta w_{\alpha,i}\Delta w_{\alpha,j}
\langle
\delta_i(\theta),\delta_j(\theta)
\rangle
\nonumber\\
&=
\Delta w_\alpha^\top
G_{\mathrm{stable}}(\theta)
\Delta w_\alpha.
\end{align}

Since $G_{\mathrm{stable}}(\theta)$ is a Gram
matrix, it is positive semidefinite.
Applying the Rayleigh quotient bound gives
\begin{align}
\|b_\alpha(\theta)\|^2
&=
\Delta w_\alpha^\top
G_{\mathrm{stable}}(\theta)
\Delta w_\alpha
\nonumber\\
&\leq
\lambda_{\max}
\bigl(G_{\mathrm{stable}}(\theta)\bigr)
\|\Delta w_\alpha\|_2^2.
\end{align}
This establishes ~\eqref{eq:systematic-bias}
and completes the proof of Lemma~\ref{lemma:bias}.
\end{proof}

\subsection{Proof of Lemma 2}

\textbf{Reference potential assumption.}
The expected full-horizon update direction
$h_0(\theta)$ is defined by averaging the OPD
surrogate gradients over trajectories sampled
from the current student policy, thereby
preserving the on-policy nature of training.
For the stationarity analysis, we additionally
assume the existence of an $L$-smooth,
lower-bounded reference potential $F$ satisfying
$\nabla F(\theta)=h_0(\theta)$.
This assumption is not required for the
reweighting-bias characterization in Lemma~\ref{lemma:bias}
and is not automatically guaranteed by the
detached OPD surrogate.
Under this assumption, Lemma~\ref{prop:on-policy-bound} characterizes
how horizon-induced reweighting bias and
stochastic gradient variance jointly affect
stationarity under an idealized stochastic
gradient update model.

\begin{proof}

Recall the update rule
\[
\theta_{t+1}
=
\theta_t-\eta\widehat g_t^\alpha.
\]
Under the assumptions of
Lemma~\ref{prop:on-policy-bound},
the conditional gradient decomposition is
\begin{equation}
\widehat g_t^\alpha
=
\nabla F(\theta_t)
+
b_\alpha(\theta_t)
+
\xi_t^\alpha,
\label{eq:app-gradient-decomposition}
\end{equation}
where
\[
\mathbb{E}[
\xi_t^\alpha\mid\mathcal{F}_t
]=0
\]
and
\[
\mathbb{E}[
\|\xi_t^\alpha\|^2\mid\mathcal{F}_t
]
\le\sigma_\alpha^2.
\]

For brevity, write
\[
u_t:=\nabla F(\theta_t),
\qquad
b_t:=b_\alpha(\theta_t),
\]
and denote conditional expectation given
$\mathcal{F}_t$ by $\mathbb{E}_t[\cdot]$.

\textbf{Step 1: One-step descent.}

By the $L$-smoothness of $F$,
\begin{align}
F(\theta_{t+1})
\le{}&
F(\theta_t)
+
\langle\nabla F(\theta_t),
\theta_{t+1}-\theta_t\rangle
\nonumber\\
&+
\frac{L}{2}
\|\theta_{t+1}-\theta_t\|^2.
\end{align}
Substituting the update rule gives
\begin{equation}
F(\theta_{t+1})
\le
F(\theta_t)
-\eta\langle u_t,\widehat g_t^\alpha\rangle
+\frac{L\eta^2}{2}
\|\widehat g_t^\alpha\|^2.
\label{eq:app-one-step}
\end{equation}

Conditioned on $\mathcal{F}_t$, both $u_t$
and $b_t$ are fixed, while
$\xi_t^\alpha$ has zero conditional mean.
Consequently,
\begin{equation}
\mathbb{E}_t[\widehat g_t^\alpha]
=
u_t+b_t,
\label{eq:app-conditional-mean}
\end{equation}
and
\begin{align}
\mathbb{E}_t
\|\widehat g_t^\alpha\|^2
&=
\mathbb{E}_t
\|u_t+b_t+\xi_t^\alpha\|^2
\nonumber\\
&=
\|u_t+b_t\|^2
+
\mathbb{E}_t\|\xi_t^\alpha\|^2
\nonumber\\
&\le
\|u_t+b_t\|^2+\sigma_\alpha^2.
\label{eq:app-second-moment}
\end{align}

Taking conditional expectation in
\eqref{eq:app-one-step} yields
\begin{align}
\mathbb{E}_t[F(\theta_{t+1})]
\le{}&
F(\theta_t)
-\eta\langle u_t,u_t+b_t\rangle
\nonumber\\
&+
\frac{L\eta^2}{2}
\left(
\|u_t+b_t\|^2+\sigma_\alpha^2
\right).
\label{eq:app-conditional-descent}
\end{align}

Since $\eta\le 1/L$, we have
$L\eta^2\le\eta$. Therefore,
\begin{align}
&-\eta\langle u_t,u_t+b_t\rangle
+\frac{L\eta^2}{2}\|u_t+b_t\|^2
\nonumber\\
&\quad\le
-\eta\langle u_t,u_t+b_t\rangle
+\frac{\eta}{2}\|u_t+b_t\|^2
\nonumber\\
&\quad=
-\frac{\eta}{2}\|u_t\|^2
+\frac{\eta}{2}\|b_t\|^2.
\label{eq:app-descent-identity}
\end{align}

Combining \ref{eq:app-conditional-descent}
and \ref{eq:app-descent-identity}
establishes
\begin{equation}
\boxed{
\begin{aligned}
\mathbb{E}_t[F(\theta_{t+1})]
\le{}&
F(\theta_t)
-\frac{\eta}{2}\|u_t\|^2\\
&+\frac{\eta}{2}\|b_t\|^2
+\frac{L\eta^2}{2}\sigma_\alpha^2.
\end{aligned}
}
\label{eq:app-descent-final}
\end{equation}

\textbf{Step 2: Telescoping over the training trajectory.}

Taking total expectation of
\eqref{eq:app-descent-final},
rearranging, and summing over
$t=1,\ldots,T$ gives
\begin{align}
\frac{\eta}{2}
\sum_{t=1}^{T}
\mathbb{E}\|u_t\|^2
\le{}&
\sum_{t=1}^{T}
\left(
\mathbb{E}F(\theta_t)
-
\mathbb{E}F(\theta_{t+1})
\right)
\nonumber\\
&+
\frac{\eta}{2}
\sum_{t=1}^{T}\mathbb{E}\|b_t\|^2
+
\frac{L\eta^2T}{2}\sigma_\alpha^2.
\end{align}

The first sum telescopes:
\begin{align}
\sum_{t=1}^{T}
\left(
\mathbb{E}F(\theta_t)
-
\mathbb{E}F(\theta_{t+1})
\right)
=
F(\theta_1)
-
\mathbb{E}F(\theta_{T+1}).
\end{align}

Using $F(\theta_{T+1})\ge F_{\inf}$,
we obtain
\begin{align}
\frac{\eta}{2}
\sum_{t=1}^{T}
\mathbb{E}\|\nabla F(\theta_t)\|^2
\le{}&
\Delta_F
+
\frac{\eta}{2}
\sum_{t=1}^{T}
\mathbb{E}\|b_\alpha(\theta_t)\|^2
\nonumber\\
&+
\frac{L\eta^2T}{2}\sigma_\alpha^2.
\end{align}

Dividing both sides by $\eta T/2$
yields
\begin{equation}
\frac{1}{T}
\sum_{t=1}^{T}
\mathbb{E}\|\nabla F(\theta_t)\|^2
\le
\frac{2\Delta_F}{\eta T}
+
L\eta\sigma_\alpha^2
+
\frac{1}{T}
\sum_{t=1}^{T}
\mathbb{E}\|b_\alpha(\theta_t)\|^2.
\label{eq:app-stationarity}
\end{equation}

Finally, applying
\eqref{eq:systematic-bias}
at each $\theta_t$ gives
\[
\frac{1}{T}
\sum_{t=1}^{T}
\mathbb{E}\|b_\alpha(\theta_t)\|^2
=
\mathfrak{B}_{\alpha,T}^2.
\]

Substitution into
\eqref{eq:app-stationarity}
proves Lemma~\ref{prop:on-policy-bound}.
\end{proof}

\subsection{Proof of Corollary 1}

Starting from
Lemma~\ref{prop:on-policy-bound},
for any $0<\eta\le 1/L$,
\begin{equation}
\frac{1}{T}
\sum_{t=1}^{T}
\mathbb{E}\|\nabla F(\theta_t)\|^2
\le
\frac{2\Delta_F}{\eta T}
+
L\eta\sigma_\alpha^2
+
\mathfrak{B}_{\alpha,T}^2.
\label{eq:app-rate-start}
\end{equation}

\textbf{Step 1: Balancing the optimization
and stochastic terms.}

Consider
\[
R_T(\eta)
=
\frac{2\Delta_F}{\eta T}
+
L\eta\sigma_\alpha^2.
\]
For $\sigma_\alpha^2>0$, its
unconstrained minimizer satisfies
\[
\frac{\partial R_T(\eta)}{\partial\eta}
=
-\frac{2\Delta_F}{\eta^2T}
+
L\sigma_\alpha^2
=0.
\]
Hence,
\begin{equation}
\eta^\star
=
\sqrt{
\frac{2\Delta_F}
{L\sigma_\alpha^2T}
}.
\label{eq:app-optimal-stepsize}
\end{equation}

If $\eta^\star\le 1/L$,
substitution yields
\begin{align}
R_T(\eta^\star)
&=
\frac{2\Delta_F}{\eta^\star T}
+
L\eta^\star\sigma_\alpha^2
\nonumber\\
&=
2\sqrt{
\frac{2L\Delta_F\sigma_\alpha^2}{T}
}.
\label{eq:app-balanced-rate}
\end{align}

To cover the constraint $\eta\le1/L$,
choose
\[
\eta
=
\min\left\{
\frac{1}{L},
\sqrt{
\frac{2\Delta_F}
{L\sigma_\alpha^2T}
}
\right\}.
\]
When $\sigma_\alpha^2=0$,
we set $\eta=1/L$.

If $\eta^\star>1/L$, then
$\sigma_\alpha^2<2L\Delta_F/T$.
Using $\eta=1/L$, we obtain
\[
R_T(1/L)
=
\frac{2L\Delta_F}{T}
+
\sigma_\alpha^2
\le
\frac{2L\Delta_F}{T}
+
2\sqrt{
\frac{2L\Delta_F\sigma_\alpha^2}{T}
}.
\]
Therefore, in either case,
\begin{equation}
R_T(\eta)
\le
2\sqrt{
\frac{2L\Delta_F\sigma_\alpha^2}{T}
}
+
\frac{2L\Delta_F}{T}.
\label{eq:app-uniform-rate}
\end{equation}

Combining
Eqs.~\eqref{eq:app-rate-start}
and~\eqref{eq:app-uniform-rate},
\begin{equation}
\frac{1}{T}
\sum_{t=1}^{T}
\mathbb{E}\|\nabla F(\theta_t)\|^2
\le
2\sqrt{
\frac{2L\Delta_F\sigma_\alpha^2}{T}
}
+
\frac{2L\Delta_F}{T}
+
\mathfrak{B}_{\alpha,T}^2.
\label{eq:app-pre-compute}
\end{equation}

\textbf{Step 2: Converting update count
to expected compute.}

By Definition 1,
\begin{equation}
\mathbb{E}[H_\alpha]
=
H_0\sum_{k=1}^{K}k^{-\alpha}.
\end{equation}

With a fixed rollout batch size $B$, define
\[
c_\alpha
=
B\mathbb{E}[H_\alpha].
\]
For an expected nominal token budget $C$,
the number of updates is
\[
T_\alpha
=
\left\lfloor
\frac{C}{c_\alpha}
\right\rfloor.
\]

For $C\ge 2c_\alpha$,
\[
T_\alpha
\ge
\frac{C}{2c_\alpha}.
\]
Consequently,
\[
\frac{1}{T_\alpha}
\le
\frac{2c_\alpha}{C},
\qquad
\frac{1}{\sqrt{T_\alpha}}
\le
\sqrt{\frac{2c_\alpha}{C}}.
\]

Substituting $T=T_\alpha$
into \eqref{eq:app-pre-compute}
gives
\begin{align}
\mathcal{E}_\alpha(C)
\le{}&
2\sqrt{
\frac{
2L\Delta_F\sigma_\alpha^2
}{T_\alpha}
}
+
\frac{2L\Delta_F}{T_\alpha}
+
\mathfrak{B}_{\alpha,T_\alpha}^2
\nonumber\\
\le{}&
4\sqrt{
\frac{
L\Delta_F\sigma_\alpha^2c_\alpha
}{C}
}
+
\frac{4L\Delta_Fc_\alpha}{C}
+
\mathfrak{B}_{\alpha,T_\alpha}^2.
\label{eq:app-compute-intermediate}
\end{align}

Finally, using
$c_\alpha=B\mathbb{E}[H_\alpha]$,
we obtain
\begin{equation}
\boxed{
\begin{aligned}
\mathcal{E}_\alpha(C)
\le{}&
4\sqrt{
\frac{
L\Delta_F\sigma_\alpha^2
B\mathbb{E}[H_\alpha]
}{C}
}\\
&+
\frac{
4L\Delta_F B\mathbb{E}[H_\alpha]
}{C}
+
\mathfrak{B}_{\alpha,T_\alpha}^2.
\end{aligned}
}\label{eq:app-compute-bound}
\end{equation}

\eqref{eq:app-compute-bound}
provides a finite-budget upper bound, retaining
both the $O(C^{-1/2})$ stochastic term and the
$O(C^{-1})$ optimization term.
For a compact characterization of the leading-order
compute--reweighting trade-off, we suppress the
lower-order $O(C^{-1})$ term and write
\begin{equation}
\mathcal{E}_\alpha(C)
\lesssim
\sqrt{
\frac{
\kappa\sigma_\alpha^2\mathbb{E}[H_\alpha]
}{C}
}
+
\mathfrak{B}_{\alpha,T_\alpha}^2,
\qquad
\kappa=16L\Delta_F B.
\label{eq:app-leading-order}
\end{equation}
Here, $\lesssim$ denotes a leading-order
representation rather than an exact
finite-budget inequality. $\kappa=16L\Delta_F B$ is independent
of $\alpha$ and $C$.
This recovers the simplified expression
in Corollary~\ref{cor:compute-tradeoff}.

\subsection{Comparison with Full-Horizon OPD.}\label{app:sub_compare}

We further characterize when stochastic horizons
yield a tighter compute-normalized stationarity
bound than standard full-horizon OPD.
Recall that the complete finite-budget bound
in \eqref{eq:app-compute-bound} is
\begin{equation}
\begin{aligned}
U_\alpha(C)
:={}&
4\sqrt{
\frac{
L\Delta_F B
\sigma_\alpha^2\mathbb{E}[H_\alpha]
}{C}
}\\
&+
\frac{
4L\Delta_F B\mathbb{E}[H_\alpha]
}{C}
+
\mathfrak{B}_{\alpha,T_\alpha}^2.
\end{aligned}
\label{eq:app-upper-bound}
\end{equation}

To ensure that the finite-budget bound applies
to both SH-OPD and full-horizon OPD,
we consider a common nominal compute budget
$C\ge 2BH_{\max}$.

At $\alpha=0$, the horizon distribution
concentrates on $H_{\max}$, and the
reweighting bias vanishes:
$\mathfrak{B}_{0,T_0}^2=0$.
Consequently,
\begin{equation}
U_0(C)
=
4\sqrt{
\frac{
L\Delta_F B\sigma_0^2 H_{\max}
}{C}
}
+
\frac{4L\Delta_F B H_{\max}}{C}.
\end{equation}

For any $\alpha>0$, comparing
$U_\alpha(C)$ with $U_0(C)$ shows that
$U_\alpha(C)<U_0(C)$ holds if and only if
\begin{align}
\mathfrak{B}_{\alpha,T_\alpha}^2
<&\
4\sqrt{\frac{L\Delta_F B}{C}}
\left(
\sqrt{\sigma_0^2 H_{\max}}
-
\sqrt{
\sigma_\alpha^2\mathbb{E}[H_\alpha]
}
\right)
\nonumber\\
&+
\frac{4L\Delta_F B}{C}
\left(
H_{\max}-\mathbb{E}[H_\alpha]
\right).
\label{eq:app-beneficial-condition}
\end{align}

This condition makes explicit the
compute--reweighting trade-off:
the reduction in compute-adjusted
stochastic optimization error must outweigh
the systematic bias introduced by horizon
reweighting.
In particular, shorter expected horizons
do not necessarily yield a tighter bound,
as the gradient variance and reweighting
bias may also vary with $\alpha$.
The comparison concerns theoretical upper
bounds under a common expected nominal
compute budget and does not guarantee
improved empirical performance.

\end{document}